\documentclass[Afour,sageh,times]{sagej}
\makeatletter
\newcommand{\mypm}{\mathbin{\mathpalette\@mypm\relax}}
\newcommand{\@mypm}[2]{\ooalign{%
		\raisebox{.1\height}{$#1+$}\cr
		\smash{\raisebox{-.6\height}{$#1-$}}\cr}}
\makeatother

\usepackage{moreverb,url}
\usepackage{amssymb, amsmath}
\usepackage{color}
\usepackage{bm}
\usepackage{textcomp}
\usepackage{listings}

\usepackage{algorithm}
\usepackage{algpseudocode}

\usepackage{titlesec}

\graphicspath{{./Figures/}}

\usepackage[usernames,dvipsnames]{xcolor}

\renewcommand{\algorithmiccomment}[1]{\bgroup\hfill//~#1\egroup}

\newcommand\BibTeX{{\rmfamily B\kern-.05em \textsc{i\kern-.025em b}\kern-.08em
T\kern-.1667em\lower.7ex\hbox{E}\kern-.125emX}}

\def\volumeyear{2019}

\begin{document}

\runninghead{Articulated soft robots in granular medium}

\title{Modeling the locomotion of articulated soft robots in granular medium}

\author{Yayun Du\affilnum{1}, Jacqueline Lam\affilnum{1}, Karunesh Sachanandani\affilnum{1}, and M. Khalid Jawed\affilnum{1}}

\affiliation{\affilnum{1}Department of Mechanical \& Aerospace Engineering, University of California Los Angeles, Los Angeles, California 90095 USA}

\corrauth{M. Khalid Jawed, Department of Mechanical \& Aerospace Engineering,
University of California Los Angeles,
Los Angeles, 
California 90095,
USA}

\email{khalidjm@seas.ucla.edu}

\begin{abstract}
Soft robots, in contrast to their rigid counter parts, have infinite degrees of freedom that are coupled with their interaction with the environment. We consider the locomotion of an untethered robot, in the granular medium, comprised of multiple flexible flagella that rotate about an axis by a motor. Drag from the grains causes the flagella to deform and the deformed shape generates a net forward propulsion. This external drag force depends on the shape of the flagella, while the change in flagellar shape is the result of the competition between the external loading and elastic forces. We introduce a numerical tool that couples discrete differential geometry based simulation of elastic rods - our model for flagella - and a resistive force theory based model for the drag. In parallel with simulations, we conduct experiments to quantify the propulsive speed of this class of robots. We find reasonable quantitative agreement between experiments and simulations. Owing to a rod-based kinematic representation of the robot, the simulation runs faster than real-time, and, therefore, we can use it as a design tool for this class of soft robots. We find that there is an optimal rotational speed at which maximum efficiency is achieved. Moreover, both experiments and simulations show that increasing the number of flagella decreases the speed of the robot. We also gain insight into the mechanics of granular medium - while resistive force theory can successfully describe the propulsion at low number of flagella, it fails when more flagella are added to the robot.
\end{abstract}



\maketitle

\begin{figure*}[t]
	\centering
	\includegraphics[width=1\textwidth]{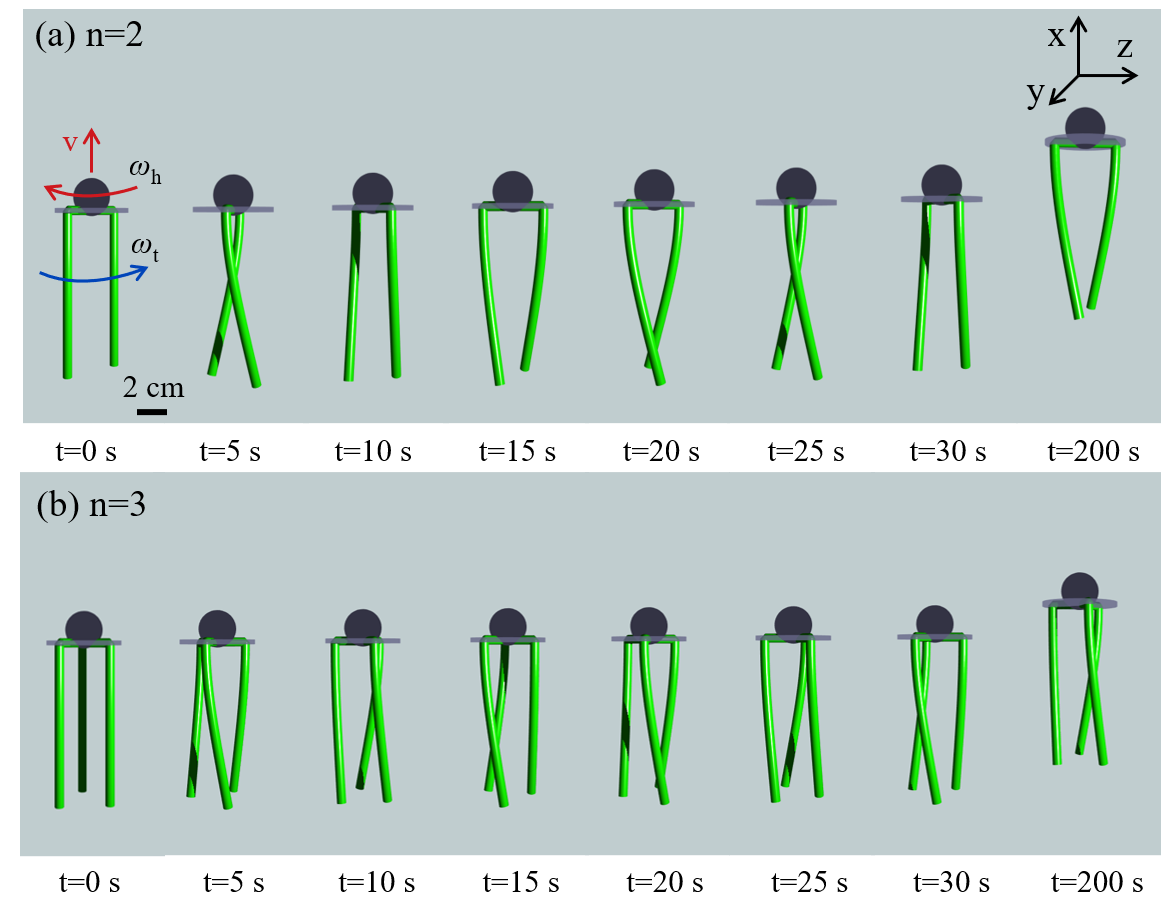}
	\caption{
	\textbf{Snapshots from simulation.}	The shape of a robot with
	(a) $n=2$ tails (Rotational speed of the motor $\omega_T=100.00$ rpm, head rotational speed $\omega_h=95.47$ rpm, tail rotational speed $\omega_t=4.53$ rpm) and
	 (b) $n=3$ tails ($\omega_T=100.00$ rpm, $\omega_h=97.49$ rpm, $\omega_t=2.51$ rpm) between $t=0$ and $t=200$ seconds. The speed of the robot is $v=0.22$ mm/s (and $v=0.13$ mm/s) for $n=2$ (and $n=3$).
}
\label{fig:intro}
\end{figure*}

\section{Introduction}
\label{sec:Intro}

Recent years have witnessed the expeditious growth of soft robots that can potentially work safely side by side with humans, reduce manufacturing costs, and revolutionize our ability to accomplish search and rescue missions~\citep{majidi2013influence}. In contrast to traditional machines and robots made of rigid skeletons or rigid materials, soft robots are primarily composed of intrinsically soft matter and fluids, enabling them to deform elastically into reversible shapes~\citep{shepherd2011multigait, lin2011goqbot, kim2013soft, tolley2014resilient, rus2015design}. This deformation can enter geometrically nonlinear regime and is often coupled with the mechanics of the surrounding medium, leading to a challenging fluid-structure interaction problem.

Nature offers a rich set of solutions to engineers seeking to design robots with locomotive ability. Depending on the mechanics of the medium, locomotion can face unique physical constraints, e.g. at fluid flow with low Reynolds number (viscous forces dominate inertia), scallop theorem states that a swimmer with time reversible motion cannot achieve propulsion~\citep{lauga2011life}. Over the past two decades, a large number of studies have been conducted on the locomotion of aquatic animals and bioinspired robots living in aquatic environment across a broad range of Reynolds number, e.g. flagellates~\citep{scaramuzza2009international, thawani2018trajectory, forghani2021control}, turtles~\citep{licht2004design}, eels \citep{yu2012bio}, fish~\citep{saimek2004motion, conte2010fast, bartlett2017high}, and octopus \citep{laschi2012soft, renda2018unified}. 

Compared with underwater locomotion, the mechanisms for underground locomotion are far less understood. This is partly due to the outstanding challenges in modeling the mechanics of granular medium and coupling it with deformable solids. Nonetheless, desert animals, such as scorpions, snakes, and lizards, show remarkable capabilities in dissipating heat from their body, facilitating feeding, reproducing, and escaping from predators. These slender flexible animals have evolved to apply a variety of locomotion modes depending on their physiology and environmental factors; most common of them are walking, running, jumping, swimming and flying \citep{gray1968animal, biewener1990biomechanics, dickinson2000animals, alexander2003principles, taylor2003flying}. A comprehensive understanding of locomotion of soft bodies in granular medium can lead to novel design of bio-inspired robots for application in hazardous terrain, e.g. search and reconnaissance through debris and underground environmental monitoring.

Fortunately, it has recently been shown that granular flow can be functionally equivalent to low Reynolds fluid flow~\citep{zhang2014effectiveness}. Flagellar propulsion, widely studied since 1955~\citep{gray1955propulsion} for application in low Reynolds fluid medium, is effective in granular medium as well~\citep{texier2017helical}. This builds a remarkable connection between the microscopic world of bacteria~\citep{lauga2009hydrodynamics} and meter-sized snakes in sand. The benefits of the flexibility of flagella, e.g. tumbling~\citep{macnab1977normal} and turning~\citep{son2013bacteria} during swimming, in viscous fluid can potentially be employed in robots for underground locomotion.

In this paper, we draw inspiration from propulsion of bacteria and introduce a palm-sized untethered robot comprised of $n \ge 2$ naturally straight elastic rods and a rigid head with embedded motor and battery. As shown in Figure~\ref{fig:intro}, the rotation of these {\em tails} brings about drag loading from the granular medium causing deformation in the soft material. As a result, the tails assume a nonlinear shape that provides a net propulsive force forward. This net propulsion is only feasible in flexible structures; in case of {\em rigid} straight tails, the propulsion is zero. We introduce a numerical method for simulation of the dynamics of a collection of Kirchhoff elastic rods~\citep{kirchhoff1859uber} under viscous drag described by Resistive Force Theory (RFT)~\citep{gray1955propulsion}. This computational tool is used to simulate the multi-limbed robot and quantitatively compared against experiments. We perform parametric studies on the speed of the robot as a function of the number of tails and rotational speed, and evaluate the optimal rotational speed for maximum efficiency. We test the applicability of RFT to granular medium and indicate regimes where this theory can fail.

\section{Background and related work}
\label{sec:background}

The problem at hand can be divided into two components: (1) the modeling of the external loading on the flexible structure of the surrounding medium  and (2) the mechanics of slender bodies composed of multiple thin elastic rods.

\subsection{Model of external loading in granular medium}
\label{sec:externalLoading}

A major challenge of modeling the nonlinear dynamics of soft robot swimming in granular media is to understand the external forces on thin filaments within sand, soil, muddy sediments and otherwise mechanically unstable terrestrial substrates that display both solid and fluid-like behavior. Desert sand is an example of these kinds of granular materials
that can display solid-like behavior in bulk and fluid-like features when disturbed. In case of purely fluid medium, modeling the motion of soft robot is always possible in principle because the rules of interaction with fluids can be worked out by solving Navier-Stokes hydrodynamics in the presence of moving boundary conditions. However, the computational cost is prohibitive for application in design and control of soft robots. For rods -- mechanical structures with one dimension much larger than the other two -- moving in low Reynolds flow, RFT is widely used to connect the hydrodynamic force from viscous environment and the velocity along the rod\rq{}s centerline~\citep{gray1955propulsion, lighthill1976flagellar, johnson1979flagellar, rodenborn2013propulsion}. Despite differences in the physical mechanisms involved, a solid friction analog to RFT in viscous fluid has been successfully applied in the context of granular media to describe the undulatory motion of sand lizards and snakes~\citep{hu2009mechanics, maladen2011mechanical}. It has also been reported that, during the movement in granular media, the primary propulsive force in slithering is generated by the anisotropic friction force of the rod surface against the substrate~\citep{hu2009mechanics} -- a mechanism reminiscent of the propulsion of bacterial flagella in viscous fluid medium. Since then, several studies have shown that the frictional forces occurring perpendicular to the body per unit length are greater than those along the body~\citep{maladen2009undulatory, ding2012mechanics, texier2017helical}.

\subsection{Mechanics of branched elastic rod structures}
\label{sec:elasticRods}

The external force from granular medium can result in geometrically nonlinear deformation in the elastic rods, as shown in Figure~\ref{fig:intro}. This coupling between the structural deformation and the forces from the granular medium in the context of an articulated soft robot is yet to be addressed in the literature. Notable prior works investigated the force on thin {\em rigid} rods in viscous fluid~\citep{lighthill1976flagellar, rodenborn2013propulsion, thawani2018trajectory} or granular medium~\citep{texier2017helical, thawani2018trajectory}. In our study, we use the Discrete Elastic Rods (DER) algorithm \citep{bergou2008discrete, bergou2010discrete, jawed2018primer} to capture the nonlinear deformation of thin elastic rods in the presence of external forces. DER method was first introduced in the computer graphics community for fast simulation of the visually dramatic dynamics of hair and other filamentary structures. Previous studies combined DER method with hydrodynamic models for viscous fluid to investigate the deformation and instability of helical elastic rod -- an analog for bacterial flagellum~\citep{jawed2015propulsion, jawed2016deformation, jawed2017dynamics}. All of these studies considered only a single elastic rod that is deforming due to hydrodynamic forces. More recently, we studied a model uniflagellar bacteria (a helical elastic rod attached to a rigid spherical head) and considered the interaction between the flows induced by the rod and the head~\citep{forghani2021control}.

A wide variety of soft robots can be modeled as a network of multiple elastic rods, optionally connected to rigid bodies. Structures comprised of multiple elastic rods, e.g. elastic gridshell (also known as Cosserat net)~\citep{baek2018form} and flexible rod mesh~\cite{perez2015design}, have also been modeled using DER. The multi-rod gridshell simulator~\citep{baek2018form} used stiff springs at the joints between two rods to impose constraints and treated the spring forces explicitly. This requires small time step size, compared with a fully implicit approach, for numerical simulation and ignores the coupling of twisting and bending modes~\citep{perez2015design} between two rods at the joints. In this study, we present a simulation algorithm that treats all the elastic and external forces implicitly in a network of rods and accounts for the presence of a rigid head. We demonstrate that a seemingly complex robot can be kinematically represented by a network of rods; this rod-based presentation can be used to leverage the computational efficiency of cutting edge tools like DER.

This paper is organized as follows. Section {\em Experimental design} provides a detailed description of the robotic platform and the experimental setup. Next the numerical model we employ for simulating the locomotion of multi-limbed robot is introduced in Section {\em Numerical model description}. The results from simulations and experiments are presented in Section {\em Results and discussion}. Finally conclusions obtained are summarized and future research directions are suggested in Section {\em Conclusion}.

\begin{figure*}[t]
	\centering
	\includegraphics[width=0.75\textwidth]{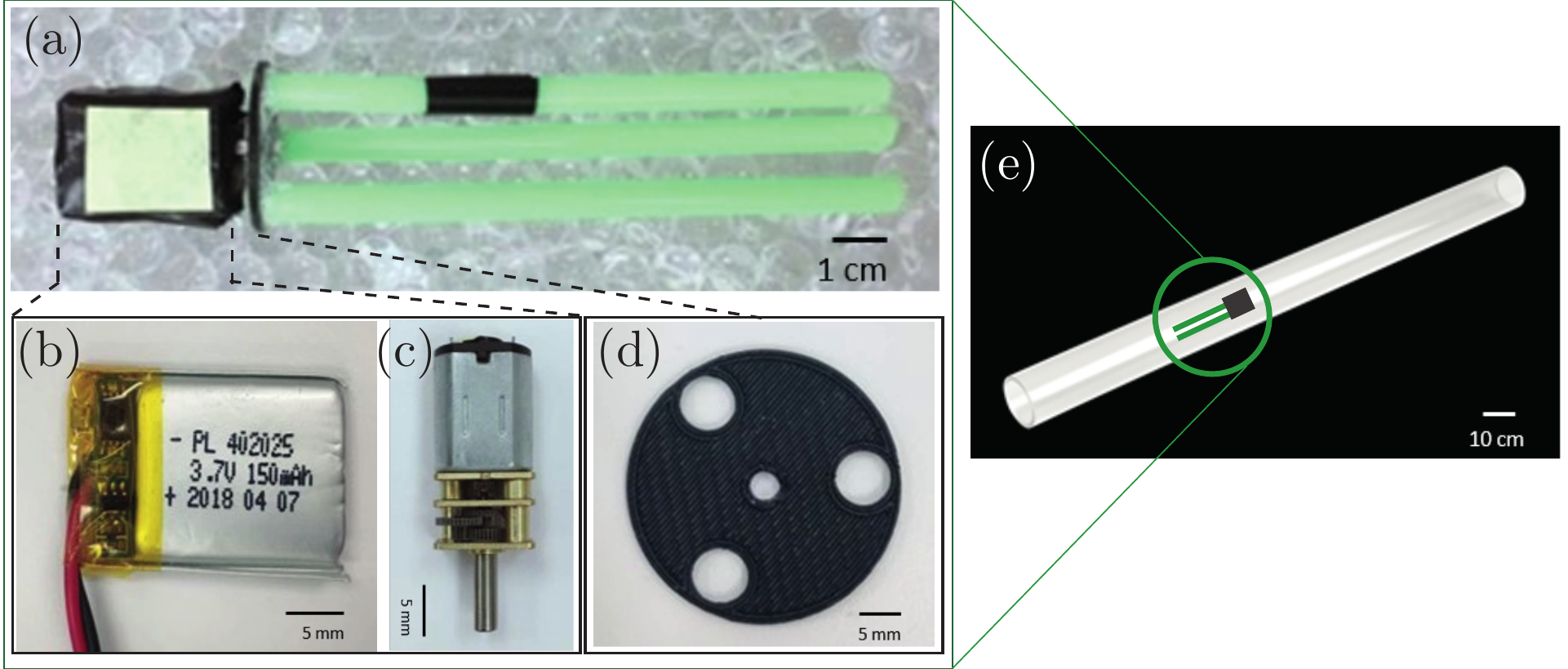}
	\caption{ The compositive view of the experimental setup. 
	(a) The robot with $n=3$ tails. The head is comprised of (b) a battery and (c) a motor. (d) A circular disc connects the tails with the head. (e) The robot is placed inside a cylindrical tube full of granular medium. 
}	
\label{fig:expt}
\end{figure*}

\section{Experimental design}
\label{sec:design}

The primary purpose of our experiments is to investigate the motion of a robot propelled by multiple elastic rods undergoing rotation. We place emphasis on the conditions and restrictions pertinent to RFT\rq{}s application to model the drag from granular media. Keeping this goal in mind, we sequentially describe the design of the robot, fabrication, experimental setup, and data analysis in the following.

\subsection{Robotic platform design} \label{subsection:robotic platform}
%
Figure~\ref{fig:expt}(a) shows a photograph of the soft robot which is a small, lightweight (14 cm, 35 g) structure actuated by $n$ number of soft elastic tails. It includes (1) a head with two 3.7V 200mAh rechargeable 502025 LiPo batteries and one DC geared motor (uxcell) with 3V nominal voltage, 0.35W nominal power and 0.55A stall current, (2) multiple elastic tails, and (3) one 3D-printed plate to hold those tails. Inside the head, two batteries are connected in parallel, naturally making the entire structure symmetric. Inspired by the head shape of desert animals such as scorpions, snakes and lizards not being perfectly spherical, we design the shape of our robot head to be cuboid to increase the ability to fluidize the granular medium in front of it. Our idea is empirically verified by the much slower movement and higher slippage occurrence of the robot with a spherical head compared to that with a cuboid head. We vary the number of robot tails to explore its effect on the translational speed, $v$, of our robot. All tails are glued to the 3D-printed plate and they are driven by a single motor through the motor shaft, protruding from the head of the robot. The control parameter is the rotational speed of the tails relative to the head, $\omega_T$. To vary this parameter, we fabricate robots with different gear motors while keeping all other components the same. The rotational speed of the gear motor is related to the voltage provided, and it will decrease when the voltage is lower. To keep the motor rotation speed consistent, we fully charge the batteries for each experimental trial and recharge the battery after every experiment that lasts for approximately 10 minutes. During data analysis, we count the number of rotations with time and find that this protocol has ensured a constant rotational speed during the entire experimental trial. Moreover, the size and weight of all the motors are almost the same, $13-15$g and $(15-17)\times 12 \times 10$mm even though they provide different rotational speeds.
When necessary, we add electrical tapes around the motor to account for the small differences in size and weight among different motors.

\subsection{Granular medium}
\label{granularMedium}
We choose water crystal bead as the granular medium to test the locomotion due to its transparency. The robot can be seen from outside the medium using a conventional digital camera (Nikon D3400). The diameter of the beads in dry state is $2.5$ mm, which increases to $d_b = 9.4 \pm 0.4$ mm after fully absorbing water. The size of beads is controllable by controlling the time they are placed inside water and changeable reversibly after dehydration. Due to this property, the water crystal beads can also be used to investigate the performance of the robot, efficiency to be mentioned in section \ref{sec:Conclusion} for example, related to the granule configuration, such as size, density and homogeneity. When performing experiments, we use the beads fully absorbing water to ensure that the size is consistent. Before experiments are carried out, we dry their surfaces  so as to decrease the occurrence possibility of slippage between the granular medium and the robot. The volume fraction -- the ratio between the solid volume and the occupied volume -- is approximately $0.52$.
The volume fraction is stated to control the response of granular media to intrusion~\citep{maladen2009undulatory} and we will discuss how it might be related to the ``stick slip'' in Section \ref{sec:Result}. It is noticeable from Figure \ref{fig:expt}(a) that the diameter of the bead is on the same order of magnitude as the diameter of the tails. RFT is intended for grains that are much smaller than the size of the robot; our choice of rather large grains is to test the limits of applicability of RFT.

\subsection{Fabrication of elastic tails}
\label{tailmaking}

The elastic tails were fabricated using a molding and casting technique developed by~\cite{lazarus2013contorting} and \cite{miller2014shapes}. A 50\% - 50\% mixture by mass of catalyst and base of a silicone-based rubber (Vinylpolysiloxane, Elite Zhermack) was injected into a PVC tube (VWR International) of inner and outer diameters are $3.175$ mm and $6.35$ mm, respectively. The PVC tube mold were affixed to a straight steel bar to hold the shape completely straight. The mixture was allowed to cure undisturbed for 24 hours.  The PVC tube was then carefully cut to extract the now solid VPS rod. The radius of elastic tails can be changed by using PVC tubes with different inner and outer diameters, making the scale-up or scale-down of our robot platform effortless.

\subsection{Locomotion experiments}
\label{locomotion}

As the reservoir of granular medium, we use a cylindrical transparent tube with with an inner radius of $53$ mm and axial length of $1220$ mm. The tube is filled with granular medium and placed horizontally (perpendicular to the direction of gravity). The robot is initially placed at one end near the center of the cross-section of the tube, meant to cancel the wall effect. Since the robot is placed at the center of the tube, surrounded by compact granules against the tube wall, the drag-induced lift mentioned in \cite{maladen2011undulatory} is suppressed. Hence, the rotation of tails will move the robot forward through the granular medium along an approximately straight line. Movements of the robot are captured by a video camera at a frame rate of $29.98$ fps. In order to count the rotational speed of the robot head ($\omega_h$) and tail ($\omega_t$), a bright yellow marker is attached to the black colored head and a black marker is put onto one of the green colored elastic tails.

\begin{figure*}[t]
\centering
\includegraphics[width=0.7\linewidth]{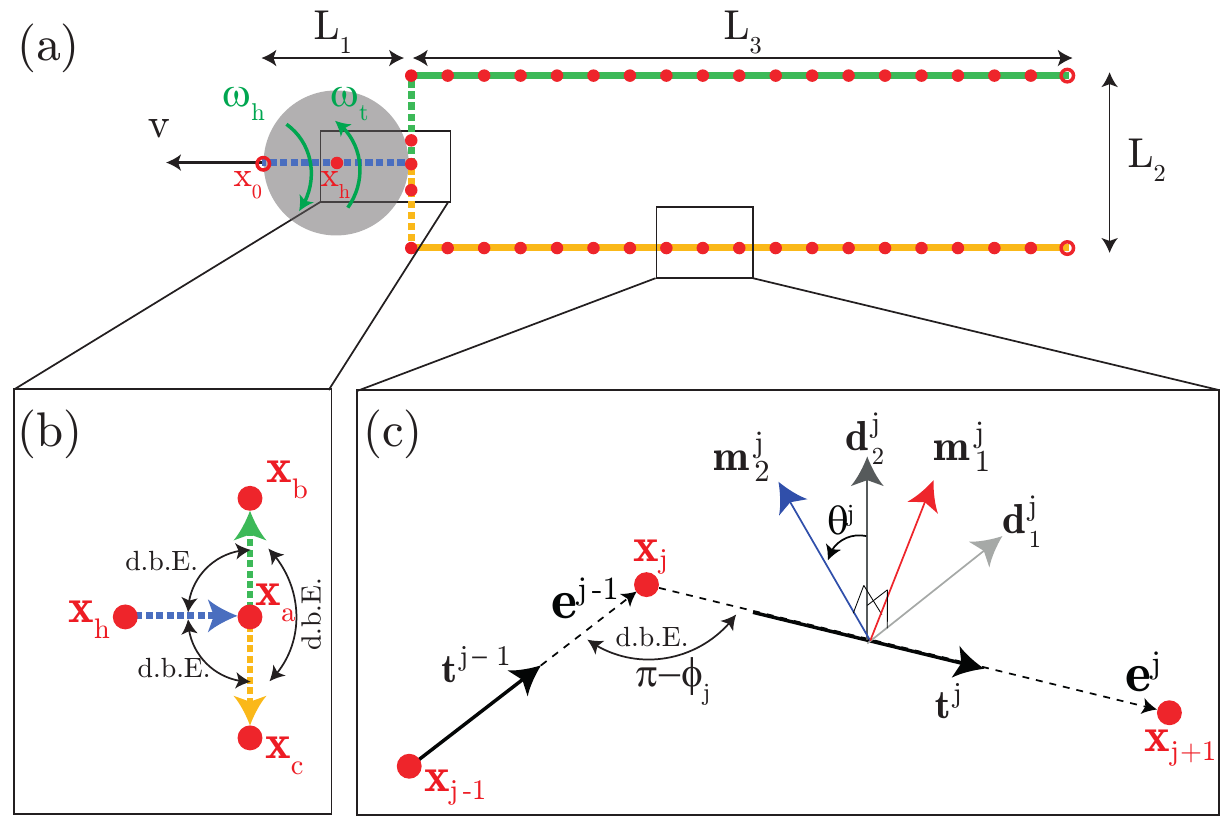}
\caption{
Schematic of the discrete representation of a robot with two tails. 
(a) Geometric parameters of the robot in undeformed state. 
Here, $L_1 = 2a$ is the diameter of the robot head, $L_2$ is the diameter of the disc connecting the head and the tails, and $L_3$ is the length of each tail. Dashed lines represent rigid structure whereas solid lines correspond to flexible structure. The node $\mathbf x_h$ represents the location of the head. (b) A close-up view of the ``joint" node $\mathbf x_a$ that connects the head with the tails. In this figure, d.B.E. indicates discrete bending and twisting energy. This is the only node that is connected to more than two nodes.
(c) A close-up view of three nodes, $\mathbf x_{j-1}, \mathbf x_j,$ and $\mathbf x_{j+1}$, and two edges, $\mathbf e^{j-1} = \mathbf x_j - \mathbf x_{j-1}$ and $\mathbf e^{j} = \mathbf x_{j+1} - \mathbf x_{j}$. The turning angle, $\phi_i$, between the two edges result in bending energy and the rotation of the material frame from one edge to the next result in twisting energy. The reference frame on $\mathbf e^j$ is $ \left\{ \mathbf{d}_{1}^{j}, \mathbf{d}_{2}^{j}, \mathbf{t}^{j} \right\} $ and the material frame is $ \left\{ \mathbf{m}_{1}^{j}, \mathbf{m}_{2}^{j}, \mathbf{t}^{j} \right\} $. The twist angle on that edge is $\theta^j$.}
\label{fig:simSchematic}
\end{figure*}

\section{Numerical model description}
\label{sec:modelingProcess}

This section introduces a numerical model of the robot where the robotic structure is represented using a network of Kirchhoff\rq{}s rods~\cite{kirchhoff1859uber}. This model is used in a numerical simulation scheme that solves the equations of motion at each degree of freedom (DOF). The simulation is subsequently utilized to extract various locomotion parameters, e.g. speed of the robot and its efficiency.

\subsection{Kinematics}
\label{sec:kinematics}

Referring to Figure~\ref{fig:simSchematic}(a), the first step in the modelling process is representing the robot as a ``stick figure". A number of nodes (circles in Figure~\ref{fig:simSchematic}) are located along the stick figure. Figure~\ref{fig:simSchematic}(b) shows the nodes at the junction between the head and the tails ($n=2$ in the figure). Node $\mathbf x_a$ is unique since it is connected to $n+1$ nodes. All the other nodes are connected to two nodes or, in case of terminal nodes (open circles in Figure~\ref{fig:simSchematic}(a)), one node. As shown in Figure~\ref{fig:simSchematic}(c), typically a node $\mathbf x_j$ is connected with two nodes $\mathbf x_{j-1}$ and $\mathbf x_{j+1}$. The vector connecting two consecutive nodes (e.g. $\mathbf e^j = \mathbf x_{j+1} - \mathbf x_j$) is called an ``edge". In addition to the location of the nodes, a complete description of the robotic structure requires a measure of the rotation of the edges. A material frame $ \left\{ \mathbf{m}_{1}^{j}, \mathbf{m}_{2}^{j}, \mathbf{t}^{j} \right\} $ ($j$ represents the edge number) is associated with each edge to keep track of rotation. This frame is orthonormal and adapted, i.e. the third director $\mathbf{t}^{j} = \mathbf{e}^{j} / \| \mathbf{e}^{j} \|$ is the tangent along the edge ($\| \cdot \|$ represents the Euclidean norm of a vector). Another orthonormal adapted frame $ \left\{ \mathbf{d}_{1}^{j}, \mathbf{d}_{2}^{j}, \mathbf{t}^{j} \right\} $ is used as the reference frame. This frame is initialized at time $t=0$ and then updated at every time step of the simulation through time-parallel transport; details will be provided later in this paper. A single scalar quantity, $\theta^j$, is necessary per edge to obtain the material frame from the reference frame as outlined in the following two equations:
\begin{subequations}
\begin{align}
\mathbf{m}_{1}^{j} &= \mathbf{d}_{1}^{j} \cos \theta^{i} + \mathbf{d}_{2}^{j} \sin \theta^{j} \label{MaterialFrame1}\\
\mathbf{m}_{2}^{j} &= - \mathbf{d}_{1}^{j} \sin \theta^{j} + \mathbf{d}_{2}^{j} \cos \theta^{j}
\label{MaterialFrame2}
\end{align}
\label{eq:materialFrame}
\end{subequations}
The angle $\theta^j$ is referred to as the ``twist angle". We follow the convention of using subscripts to denote node-based quantities and superscripts for edge-based quantities.

The locations of the nodes, $\mathbf x_j$ ($0 \le j < N$ where $N$ is the number of nodes), and the twist angles, $\theta^j$ ($0 \le j < N_e$ where $N_e$ is the number of edges), completely describe the configuration of the robot. For the robot studied in this paper, it turns out that $N_e = N - 1$ (see Figure~\ref{fig:simSchematic}(a)). The DOF vector for the robot is
\begin{equation}
\mathbf q = \left[
\mathbf x_0, \mathbf x_1, \mathbf x_2, \ldots, \mathbf x_{N-1}, \theta^0, \theta^1, \ldots, \theta^{N_e-1} \right]^T,
\end{equation}
where the superscript $^T$ denotes transpose. If a robot has $N$ nodes, the size of $\mathbf q$ is $\textrm{ndof}=3\times N + N_e$. Since the robot deforms with time, the DOF vector is a function of time, i.e. $\mathbf q \equiv \mathbf q (t)$. Knowing the configuration of the robot at $t=0$ (i.e. $\mathbf q(0)$ is known), the task at hand is to compute $\mathbf q (t)$. This is achieved by discretizing the time into small steps of step size $\Delta t$ and solving the equations of motion at each time step.

Equations of motion are statements of balance of forces. The internal forces in the robotic structure arises from the elastic nature of the material. In this paper, the rigid components of the robot (e.g. the head and the disc indicated by dashed lines in Figure~\ref{fig:simSchematic}(a)) are assumed to be elastic with high elastic stiffness so that the deformation in these components are negligible compared with the deformation in the flexible tails. In the following, the strains in the structure, the elastic energies associated with these strains, and the elastic and external forces are sequentially discussed.

\subsection{Macroscopic strains}
\label{sec:strains}

At time $t=0$, the robot is undeformed with zero strains and DOF vector is $\mathbf q(0) \equiv \bar{\mathbf q}$; hereafter, $\bar{(\; )}$ represents evaluation of a quantity in undeformed configuration. Even though the undeformed and initial configurations are the same in the system studied here, this is not a required assumption for the simulation scheme.

Axial stretch, curvature, and twist are the macroscopic strains along the structure~\cite{audoly2000elasticity}. As outlined next, these strain measures can be computed from the DOF vector $\mathbf q$ and are used to compute the elastic energy.

Axial stretch is an edge-based quantity that is related to the elongation of an edge. The axial stretch, $\epsilon^{j}$, in the $j$-th edge is
\begin{equation}
\epsilon^{j} = \frac { \| \mathbf{e}^{j} \| } { \| \bar{\mathbf{e}}^{j} \| } - 1.
\label{eq:stretch}
\end{equation}

Curvature is a node-based quantity that is related to the turning angle $\phi_j$ (see Figure~\ref{fig:simSchematic}(c)). No curvature is associated with the terminal nodes. Curvature binormal is a vector representing the turn:
\begin{equation}
(\mathbf{\kappa b})_{j} = \frac {2 \mathbf{e}^{j-1} \times \mathbf{e}^{j} } { \| \mathbf{e}^{j-1} \| \| \mathbf{e}^{j} \| + \mathbf{e}^{j-1} \cdot \mathbf{e}^{j} }.
\end{equation}
It turns out that $\| (\mathbf{\kappa b})_{j} \|  = 2 \tan \left( \frac{\phi_j}{2} \right)$. The curvature of the osculating circle passing through $\mathbf x_{j-1}, \mathbf x_j$, and $\mathbf x_{j+1}$ is $\| (\mathbf{\kappa b})_{j} \| / \Delta l$ where $\Delta l = \| \mathbf{e}^{j} \| = \| \mathbf{e}^{j-1} \|$. The scalar curvatures along the first and second material directors are
\begin{subequations}
\begin{align}
\kappa_{j}^{(1)} &= \frac {1} {2} (\mathbf{m}_{2}^{j-1} + \mathbf{m}_{2}^{j}) \cdot (\mathbf{\kappa b})_{j}, \label{BendingCurvature1} \\
\kappa_{j}^{(2)} &= \frac {1} {2} (\mathbf{m}_{1}^{j-1} + \mathbf{m}_{1}^{j}) \cdot (\mathbf{\kappa b})_{j} \label{BendingCurvature2}.
\end{align}
\end{subequations}

Associated with every curvature is a twist that represents the rotation of the material frame from one edge to the next edge. In Figure~\ref{fig:simSchematic}(c), the twist at the $j$-th node is
\begin{equation}
\tau_{j} = \theta^j - \theta^{j-1} + \Delta m_{j, \textrm{ref}},
\end{equation}
where $\Delta m_{j, \textrm{ref}}$ is the reference twist that represents the twist of the reference frame as it moves from the $(j-1)$-th edge to the $j$-th edge. The procedure to calculating this reference twist is discussed next. The first director of the reference frame, $ \mathbf{d}_{1}^{j-1}$, is {\em parallel transported} from the $(j-1)$-th edge to the $j$-th edge to get $\mathbf{d}_\textrm{tmp}$. Parallel transport is the process of moving the reference director from one edge to the next without twist; it involves the following steps.
\begin{eqnarray*}
& \mathbf b &= \mathbf t^{j-1} \times \mathbf t^j,\\
& \hat {\mathbf b} &= \frac{\mathbf b}{|\mathbf b|},\\
& \mathbf n_1 &= \mathbf t^{j-1} \times \hat {\mathbf b},\\
& \mathbf n_2 &= \mathbf t^j \times \hat {\mathbf b},\\
& \mathbf{d}_\textrm{tmp} &= (\mathbf{d}_{1}^{j-1} \cdot \mathbf t^{j-1}) \mathbf t^j + (\mathbf{d}_{1}^{j-1} \cdot \mathbf n_1) \mathbf n_2 + (\mathbf{d}_{1}^{j-1} \cdot \hat {\mathbf b}) \hat {\mathbf b},
\end{eqnarray*}
where $\mathbf t^{j-1}$ and $\mathbf t^j$ are the tangents on the $(j-1)$-th and $j$-th edges, respectively. The reference twist, $\Delta m_{j, \textrm{ref}}$, is the signed angle from $\mathbf{d}_\textrm{tmp}$ to $\mathbf{d}_{1}^{j}$ about $\mathbf t^j$.

\subsection{Elastic energies}
\label{sec:energies}

The total elastic energy of the structure is the linear sum of stretching, bending, and twisting energies such that
\begin{equation}
E_\textrm{elastic} = E_s + E_b + E_t,
\label{eq:elasticEnergy}
\end{equation}
where $E_s, E_b,$ and $E_t$ are the stretching, bending, and twisting energies, respectively. 

Associated with each edge is a discrete stretching energy that can be computed from the axial stretch in Eq.~\ref{eq:stretch},
\begin{eqnarray}
E_s = \sum \frac{1}{2} EA \left( \epsilon^{j} \right)^2 \| \bar{\mathbf{e}}^{j} \|,
\label{eq:stretchingEnergy}
\end{eqnarray}
where $\sum$ represents summation over all the edges, $E$ is the Young\rq{}s modulus, $A=\pi r_0^2$ is the cross-sectional area, $r_0$ is the cross-sectional radius. For the edges that are located on rigid part of the robot (head and disc), the parameter $EA$ is taken to be sufficiently large so that the deformation is negligible. 

The bending energy is
\begin{equation}
E_{\text{b}} = \sum \frac {1} {2} \frac {EI} {\Delta l_{j}} \left[ \left( \kappa_{j}^{(1)} - \bar{\kappa}_{j}^{(1)} \right)^2 + \left( \kappa_{j}^{(2)} - \bar{\kappa}_{j}^{(2)} \right)^2 \right],
\label{eq:bendingEnergy}
\end{equation}
where $\sum$ represents summation over all the curvatures, $ \Delta l_{j} = \frac {1}{2} \left( \| \bar{\mathbf{e}}^{j-1} \| + \| \bar{\mathbf{e}^{j}} \| \right) $ is the Voronoi length associated with the $j$-th node, $ \bar{\kappa}_{j}^{(1)} $ and $ \bar{\kappa}_{j}^{(2)} $ are the material curvatures in undeformed configuration, and $EI = \frac{\pi}{4}Er_0^4$ is the bending stiffness. To model rigid components of the robot, the bending stiffness is assumed to be large enough so that the curvatures at the rigid nodes remain almost constant throughout the simulation. 

The twisting energy is
\begin{equation}
E_t = \sum \frac {1} {2} \frac {GJ} {\Delta l_{j}} \left( \tau_{j} - \bar{\tau_{j}} \right)^2,
\label{eq:twistingEnergy}
\end{equation}
where $ \bar{\tau_{j}} $ is the undeformed twist along the centerline, $G$ is the shear modulus, and $GJ =  \frac{\pi}{2}Gr_0^2$ is the twisting stiffness. This stiffness is assumed to be sufficiently large for the rigid components. The material of the tail is nearly incompressible (i.e. Poisson\rq{}s ratio $\nu=0.5$) and therefore $G = E/3$.

In case of a single elastic rod, each internal node is associated with a discrete bending energy and a discrete twisting energy. However, the robot is represented as a network of rods and the ``joint" node ($\mathbf x_a$ in Figure~\ref{fig:simSchematic}(b)) has multiple discrete bending energies (indicated by d.B.E. in Figure~\ref{fig:simSchematic}(b)) and discrete twisting energies associated with it. This observation is important during the programming implementation of the simulation algorithm.

\subsection{External forces using Resistive Force Theory}
\label{sec:externalForces}

\begin{figure}[h]
\centering
\includegraphics[width=\linewidth]{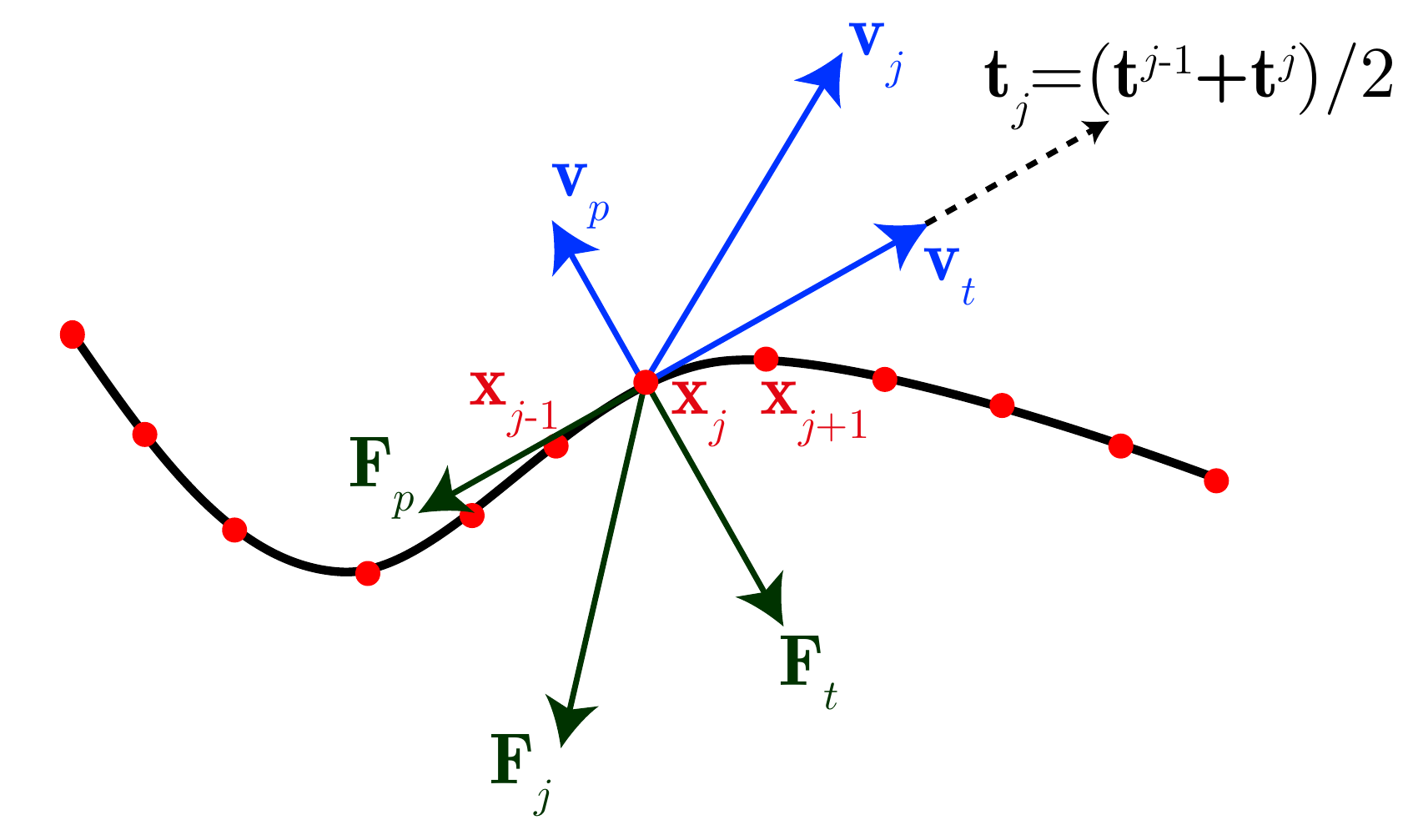}
\caption{Schematic representation of RFT.}
\label{fig:RFT}
\end{figure}

First, the external force on the tails exerted by the granular medium following RFT is discussed. In Figure \ref{fig:RFT}, we schematically represent a slender rod in discrete setting moving in the granular medium. The velocity, $\mathbf v_j \equiv \dot{\mathbf x}_j$ at point $ \mathbf{x}_{j} $ can be decomposed into two parts: the parallel term $ \mathbf{v}_t = (\mathbf v_j \cdot \mathbf t_j) \mathbf t_j$ and the perpendicular term $ \mathbf{v}_p = \mathbf{v}_j - \mathbf{v}_t$, where the tangent at the $j$-th node $\mathbf t_j = \frac{1}{2} (\mathbf t^{j-1} + \mathbf t^j)$ is the average of the tangents along the two associated edges. The tangential and perpendicular forces from the medium that resist $ \mathbf{v}_t$ and $ \mathbf{v}_p$ are
\begin{subequations}
\begin{align}
\mathbf F_t &= - \eta_{t} \mathbf v_t \Delta l_j,\\
\mathbf F_p &= - \eta_{p} \mathbf v_p \Delta l_j,
\end{align}
\label{eq:externalForceTail}
\end{subequations}
where the drag coefficients along the tangential and perpendicular directions~\citep{gray1955propulsion} are
\begin{subequations}
\begin{align}
\eta_{t} &= 2 \pi \mu /\left[ \log(\frac {2L} {r_0}) - \frac {1} {2} \right] \\
\eta_{p} &= 4 \pi \mu /\left[ \log(\frac {2L} {r_0}) + \frac {1} {2} \right], 
\end{align}
\end{subequations}
$ \mu $ is the dynamic viscosity, and $ L $ is the length of tail ($ L = L_3 $ in Figure~\ref{fig:simSchematic}(a)).
The external force on the $j$-th node, if this node is located on the tails, is
\begin{equation}
    \mathbf F_j = \mathbf F_t + \mathbf F_p.
\end{equation}

In addition to the tails, the head of the robot is also rotating and translating. The rotational speed of the head ($\omega_h$ in Figure~\ref{fig:simSchematic}) can be extracted from the time derivative of the twist angle, $\theta^h$, of the edge connected $\mathbf x_0$ and $\mathbf x_h$, i.e. $\omega_h \equiv \dot{\theta}^h$. The velocity of the head, $\mathbf v_h$ is the time derivative of $\mathbf x_h$, i.e. $\mathbf v_h \equiv \dot{\mathbf x}_h$. If the head was spherical with radius $a$, the viscous drag according to Stokes\rq{}s law would result in an external force on $\mathbf x_h$,
\begin{equation}
\textbf{F}_h = - 6\pi \mu a \mathbf v_h,
\label{eq:stokes_drag}
\end{equation}
and an external torque on the edge (with twist angle $\theta^h$),
\begin{equation}
T_h = - 8\pi \mu a^3 \omega_h.
\label{eq:stokes_torque}
\end{equation}
In our case, the robot head shape is not a sphere and we cannot directly apply Eqs. \ref{eq:stokes_drag} and \ref{eq:stokes_torque}. Instead, we use a numerical coefficient, $C_1$, to account for the shape of the robot head. As a result, Eq. \ref{eq:stokes_drag} is updated as follows
\begin{equation}
\textbf{F}_h = - (6\pi C_1) \mu a \mathbf v_h.
\label{eq:stokes_drag_refined}
\end{equation}
Similarly, Eq. \ref{eq:stokes_torque} is updated to include a numerical coefficient, $C_2$, that accounts for the nonspherical shape of the head:
\begin{equation}
T_h = - (8\pi C_2) \mu a^3 \omega_h .
\label{eq:stokes_torque_refined}
\end{equation}
The coefficients $C_1$ and $C_2$ are used as fitting parameters in Section \ref{subsection:parameterfitting} and obtained through data fitting. 

\subsection{Simulation loop, equations of motion}
\label{sec:simulationLoop}

In the simulation scheme (Algorithm~\ref{algo:discreteElasticRods}), time is discretized into small time-steps and the configuration of the robot represented by the DOF vector, $\mathbf q$, is updated at each time step. The equation of motion at the $i$-th DOF to march from $t=t_k$ to $t=t_{k+1}=t_k + \Delta t$ ($\Delta t$ is the time step size) is
\begin{equation}
f_i \equiv \frac{m_i} {\Delta t} \left[
\frac{ q_i (t_{k+1}) - q_i (t_k) } { \Delta t } - 
\dot{q}_i (t_k) \right] +
\frac{\partial E_{\textrm{elastic}}}{\partial q_i} -
f_i^{\textrm{ext}} = 0,
\label{eq:DER3D_EOM}
\end{equation}
where $i=1, \ldots, \mathrm{ndof}$, the {\em old} DOF $q_i (t_k)$ and velocity $\dot{q}_i (t_k)$ are known, $E_{\textrm{elastic}}$ is the elastic energy evaluated at $q_i (t_{k+1})$, $f_i^{\textrm{ext}}$ is the external force (or moment for twist angles) on the $i$-th DOF, and $m_i$ is the lumped mass at each DOF. Since the dynamics of the system is dominated by viscosity with negligible influence of inertia, the results presented in this paper do not vary with the mass parameters as long as low Reynolds number is maintained. Note that Eq.~\ref{eq:DER3D_EOM} is simply a statement of ``mass times acceleration = elastic force + external force" at the $i$-th DOF. Eq.~\ref{eq:DER3D_EOM} represents a system of $\textrm{ndof}$ equations that has to be solved to obtain the {\em new} DOF $q_i (t_{k+1})$. Once the new DOF is obtain, the new velocity is simply $\dot{q}_i (t_{k+1}) = \left( q_i(t_{k+1}) - q_i(t_{k}) \right) / \Delta t$.

Newton-Raphson method is used to solve the equations of motion. Referring to Algorithm~\ref{algo:discreteElasticRods}, this involves solving the following linear system of size $\textrm{ndof}$,
\begin{equation}
    \mathbb J \Delta \mathbf q = \mathbf f,
\end{equation}
where $\mathbf f$ is a vector of size $\textrm{ndof}$, the $i$-th component of this vector can be computed from Eq.~\ref{eq:DER3D_EOM}, and $\mathbb J$ is a square matrix representing the Jacobian for Eq.~\ref{eq:DER3D_EOM}. The $(i,j)$-th component of the Jacobian is
\begin{equation}
\mathbb J_{ij} = \frac{\partial f_i}{\partial \xi_j} = \mathbb J^{\textrm{inertia}}_{ij} + \mathbb J^{\textrm{elastic}}_{ij} + \mathbb J^{\textrm{ext}}_{ij},
\label{eq:DER3D_Jacobian}
\end{equation}
where
\begin{eqnarray}
\mathbb J^{\textrm{inertia}}_{ij} = \frac{m_i}{\Delta t^2} \delta_{ij},\\
\mathbb J^{\textrm{elastic}}_{ij} = \frac{\partial^2 E_{\textrm{elastic}}}{\partial q_i \partial q_j},\\
\mathbb J^{\textrm{ext}}_{ij} = - \frac{\partial f_i^{\textrm{ext}}} {\partial q_j}.
\label{eq:externalForceJacobian}
\end{eqnarray}
Here, $\delta_{ij}$ represents Kronecker delta.
Evaluation of the gradient of the elastic energy ($\frac{\partial E_{\textrm{elastic}}}{\partial q_i}$) as well as its Hessian ($\frac{\partial^2 E_{\textrm{elastic}}}{\partial q_i \partial q_j}$) are well documented in ~\cite{jawed2018primer, bergou2010discrete}.

\begin{algorithm*}
\caption{Discrete Simulation of Robots}\label{algo:discreteElasticRods}
\begin{algorithmic}[1]
\Require{$\mathbf q (t_k), \dot{\mathbf q} (t_k)$} \Comment{DOFs and velocities at $t=t_j$}
\Require{$\left( \mathbf d_1^j (t_k), \mathbf d_2^j (t_k), \mathbf t^j (t_k) \right)$, $0 \le j < N_e$} \Comment{Reference frame at $t=t_k$}
\Ensure{$\mathbf q (t_{k+1}), \dot{\mathbf q} (t_{k+1})$} \Comment{DOFs and velocities at $t=t_{k+1}$}
\Ensure{$\left( \mathbf d_1^j (t_{k+1}), \mathbf d_2^j (t_{k+1}), \mathbf t^j (t_{k+1}) \right)$, $0 \le j < N_e$} \Comment{Reference frame at $t=t_{k+1}$}
\Statex
\Function{Discrete Simulation of Robots}{$\; \mathbf q (t_k), \dot{\mathbf q} (t_k), \left( \mathbf a_1^j (t_k), \mathbf a_2^j (t_k), \mathbf t^j (t_k) \right)\;$}
    \State{$\bar \tau_h (t_k) \gets \omega_T t_k$} \Comment{Actuation using Eq.~\ref{eq:SpecifyTheta}}
    \State {Guess: $\mathbf q^{(1)} \gets  \mathbf q (t_{k})$}
    \State {$n \gets 1$}
    \While{\texttt{error} $>$ \texttt{tolerance}} \Comment{Newton-Raphson iterations}
    	\State Compute reference frame $ 
\left( \mathbf d_1^j, \mathbf d_2^j, \mathbf t^j \right)^{(n)}$
        \Comment{Parallel transport $\mathbf d_1^j (t_{k})$ and $\mathbf d_2^j (t_{k})$ from \ldots}
        \State{}
        \Comment{\ldots $\mathbf t^j (t_{k})$  to tangent on $j$-th edge in $\mathbf q^{(n)}$ to get $\left(\mathbf d_1^j\right)^{(n)}$ and $\left(\mathbf d_2^j\right)^{(n)}$}
    	\State Compute reference twist $\Delta m_{j, \textrm{ref}}^{(n)}$ at each internal node
    	\State Compute material frame $ 
\left( \mathbf m_1^j, \mathbf m_2^j, \mathbf t^j \right)^{(n)}$     
        \Comment{Eq.~\ref{eq:materialFrame}}
    	\State Compute $\mathbf f$ and $\mathbb{J}$ \Comment{Eqs.~\ref{eq:DER3D_EOM} and \ref{eq:DER3D_Jacobian}}
    	\State $\Delta \mathbf q \gets \mathbb J \backslash \mathbf f$ 
    	\State $\mathbf q^{(n+1)} \gets \mathbf q^{(n)} - \Delta \mathbf q$ \Comment{Update DOFs}
    	\State \texttt{error} $\gets$ \texttt{ sum ( abs ( } $\mathbf f$ \texttt{) ) }
    	\State $n \gets n+1$
    \EndWhile
	\Statex
	\State $\mathbf q (t_{k+1}) \gets \mathbf q^{(n)}$
	\State $\dot{\mathbf q} (t_{k+1}) \gets \frac{ \mathbf q (t_{k+1}) - \mathbf q (t_k) } {\Delta t}$
	\State $\left( \mathbf d_1^j (t_{k+1}), \mathbf d_2^j (t_{k+1}), \mathbf t^j (t_{k+1}) \right) \gets 
\left( \mathbf d_1^j, \mathbf d_2^j, \mathbf t^j) \right)^{(n)}$
    \State \Return {$\mathbf q (t_{k+1}), \dot{\mathbf q} (t_{k+1}), \left( \mathbf d_1^j (t_{k+1}), \mathbf d_2^j (t_{k+1}), \mathbf t^j (t_{k+1}) \right)$}
\EndFunction
\end{algorithmic}
\end{algorithm*}

\subsection{Actuation of the robot}
\label{subsection:twistingActuation}

An important contribution of this study is the observation that the actuation (e.g. rotation of motor in the robot) can be readily accounted for in the above framework by updating the undeformed configurations with time. Typically, undeformed configuration of a structure is fixed and assumed to be constant through the simulation. The strains in undeformed configuration (e.g. $\bar{\kappa}_{j}^{(1)}, \bar{\kappa}_{j}^{(2)}, \bar \tau_j$) are used in calculation of elastic energies, their gradient (i.e. elastic forces), and Hessian. However, in case of this robot, the rotation of the motor causes the undeformed twist at the head node ($\mathbf x_h$) to vary with time. If the rotational speed of the motor is $\omega_T$, we assume that the undeformed twist at the head node is
\begin{equation}
\bar \tau_h (t_k) = \omega_T t_{k}.
\label{eq:SpecifyTheta}
\end{equation}
This results in rotations of the head ($\omega_h$) and the tails ($\omega_t$) along opposite directions such that $|\omega_T| = |\omega_h| + |\omega_t|$. The total rotational speed, $\omega_T$, is a control parameter in this study.

\begin{figure*}[t]
\centering
\includegraphics[width=1.00\textwidth]{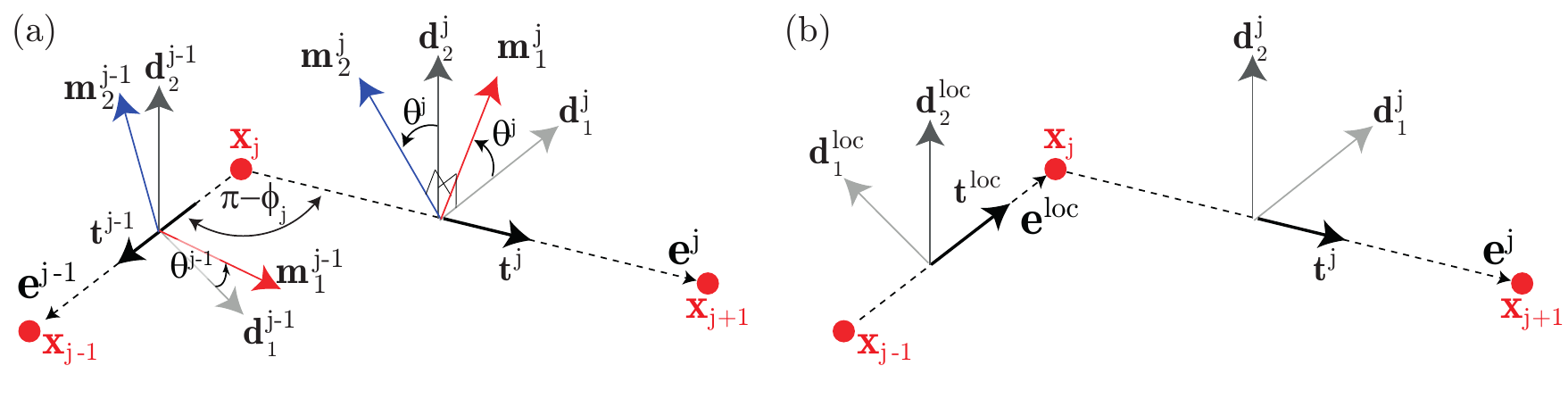}
\caption{(a) Discrete bending and twisting energy is located at $\mathbf x_j$. Both the edges are pointing away from $\mathbf x_j$.
(b) One of the edges ($\mathbf e^{j-1}$ in this case) is flipped to slightly modify the kinematic representation. This representation can be used to compute the gradient and Hessian of the bending and twisting energies following analytical expressions available in the literature~\citep{bergou2010discrete}.
}
\label{fig:sign}
\end{figure*}

\subsection{Remarks on implementation of the algorithm}
\label{sec:implementationNotes}

The most computationally expensive part of Algorithm~\ref{algo:discreteElasticRods} is solving the linear system (Line 11). To reduce computation time, it is important to observe the sparsity of the Jacobian matrix, $\mathbb J$, and exploit this sparsity during the solution process~\citep{schenk2002solving}. Referring to Figure~\ref{fig:simSchematic}(c), the entire structure can be assumed to be a combination of stretching springs (e.g. one stretching spring is between $\mathbf x_j$ and $\mathbf x_{j+1}$) and bending-twisting springs (e.g. one bending-twisting spring is between $\mathbf x_{j-1}, \mathbf x_j,$ and $\mathbf x_{j+1}$). The stretching energy of each spring (Eq.~\ref{eq:stretchingEnergy}) depends only on six DOFs (nodal coordinates of two nodes). For the stretching spring on edge $\mathbf e^j$, these DOFs are $\mathbf x_j$ and $\mathbf x_{j+1}$. The gradient vector $\left( \frac{\partial}{\partial \mathbf q} \left[ \frac{1}{2} EA \left( \epsilon^{j} \right)^2 \| \bar{\mathbf{e}}^{j} \| \right] \right)$ has only six non-zero terms and the Hessian matrix $\left( \frac{\partial^2}{\partial \mathbf q \partial \mathbf q} \left[ \frac{1}{2} EA \left( \epsilon^{j} \right)^2 \| \bar{\mathbf{e}}^{j} \| \right] \right)$ has only $6 \times 6$ non-zero terms.
The bending energy and the twisting energy of each spring (Eqs.~\ref{eq:bendingEnergy} - \ref{eq:twistingEnergy}) depend only on eleven DOFs, i.e. $\mathbf x_{j-1}, \theta^{j-1}, \mathbf x_j, \theta^j,$ and $\mathbf x_{j+1}$ in case of the spring located at $\mathbf x_j$ in Figure~\ref{fig:simSchematic}(c). The gradient vector and the Hessian matrix of these two energies therefore have only eleven and $11 \times 11$ non-zero terms. The full expressions for the gradient and Hessian terms can be found in ~\cite{bergou2008discrete}, ~\cite{jawed2018primer}, and ~\cite{panetta2019x}; software implementation is also available in open-source repositories accompanying ~\cite{jawed2014coiling, panetta2019x, choi2021implicit}.

The specific simulation studied in this paper requires the gradient of the external forces (Eq.~\ref{eq:externalForceJacobian}). The external forces are expressed in Eqs.~\ref{eq:externalForceTail},~\ref{eq:stokes_drag_refined}, and ~\ref{eq:stokes_torque_refined}. Their gradients with respect to the DOFs can be trivially obtained. Note that $\mathbb J_{ij}^\textrm{ext}$ is sparse. Since the expressions of all the Jacobian terms can be analytically evaluated and incorporated into the software, the simulation can use Euler-backward method. In comparison with Euler-forward method, Euler-backward method typically can converge at larger values of $\Delta t$ and requires less computation time.

If the structure to be simulated is a single elastic rod (unlike a network of rods in this paper), the Jacobian is a banded matrix~\cite{bergou2010discrete}. In this paper, the Jacobian is not banded due to the presence of the joint node $\mathbf x_a$ in Figure~\ref{fig:simSchematic}(b). A second difference is related to the implementation of the gradient and Hessian of the bending and twisting energies. As in Figure~\ref{fig:simSchematic}, the expressions for gradient and Hessian in ~\cite{bergou2010discrete} assume that the tangent $\mathbf t^{j-1}$ is pointing towards $\mathbf x_j$ and the second tangent $\mathbf t^{j}$ is pointing award from $\mathbf x_j$. Since this paper studies a network of rods, this assumption does not always hold. For example, as represented in Figure~\ref{fig:sign}(a), we can have cases where both the tangents ($\mathbf t^{j-1}$ and $\mathbf t^{j}$) point away from $\mathbf x_j$, the location of the bending and twisting spring. In this case, we can simply flip the first tangent ($\mathbf t^\textrm{loc} = - \mathbf t^{j-1}$ in Figure~\ref{fig:sign}) and use $ \left\{ \mathbf{d}_{1}^\textrm{loc} = - \mathbf{d}_{1}^{j-1}, \mathbf{d}_{2}^\textrm{loc} = \mathbf{d}_{2}^{j}, \mathbf t^\textrm{loc} = - \mathbf t^{j-1} \right\} $ as the ``local" reference frame on the edge $\mathbf e^\textrm{loc} = \mathbf x_j - \mathbf x_{j-1}$. The reference frame on the other edge $\mathbf e^j$ remains unchanged to $ \left\{ \mathbf{d}_{1}^{j}, \mathbf{d}_{2}^{j}, \mathbf{t}^{j} \right\} $. Flipping the edge also implies that the twist angle on $\mathbf e^\textrm{loc}$ in this local representation is $\theta^\textrm{loc} = - \theta^{j-1}$. This local representation in Figure~\ref{fig:simSchematic}(b) can be used to compute the gradient and Hessian of the bending and twisting energies at $\mathbf x_j$ with respect to $\left\{ \mathbf x_{j-1}, \theta^\textrm{loc}, \mathbf x_j, \theta^j, \mathbf x_{j+1} \right\}$ following the analytical expressions available in \cite{bergou2010discrete}. Prior to including these gradient and Hessian terms in $\mathbf f$ (Eq.~\ref{eq:DER3D_EOM}) and $\mathbb J$ (Eq.~\ref{eq:DER3D_Jacobian}), we have to be mindful that $\frac{\partial}{\partial \theta^{j-1}} (\,) = - \frac{\partial}{\partial \theta^\textrm{loc}} (\,)$.

\subsection{Physical parameters}
\label{sec:physicalParameters}

The material and geometric parameters of the robot during experiments are listed as follows:
Young's modulus $E = 1.2 \times 10^6$ N/m$^2$, Poisson\rq{}s ratio $\nu=0.5$, density of soft tails $1000$kg/m$^3$ (this is used to compute $m_i$ in Eq.~\ref{eq:DER3D_EOM}), and cross-sectional radius of tails $r_0=3.2$ mm. 
The length of each flagellum is $L_3=0.111$ m, radius of the robot head is $a=0.02$ m, and the diameter of 3D-printed circular disc is $L_2=0.04$ m.
For the simulation data presented in this paper, time step is $\Delta t=10^{-2}$ s and the length of each edge on tails (in undeformed state) is $\| \bar {\mathbf e}^j \| = 4.11$ mm. We performed convergence studies to ensure that the size of temporal and spatial discretization ($\Delta t, \| \bar {\mathbf e}^j \|$) has negligible effect on the simulation results. The parameters $\mu, C_1$, and $C_2$ will be fitted later in Section~\ref{subsection:parameterfitting}.

\begin{figure*}[t]
	\centering
	\includegraphics[width=0.9\textwidth]{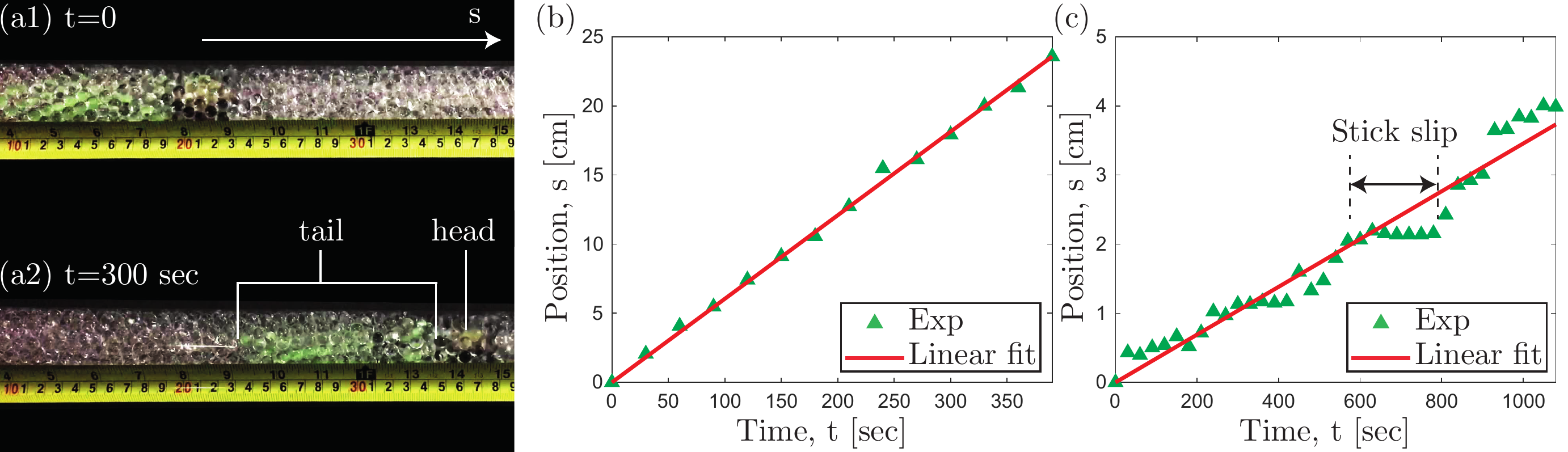}
	\caption{
		\textbf{Position of the robot with time.}	
		(a1-a2) Snapshots from experiments showing the location of a robot with $n=2$ and $\omega_{\textrm{total}}=250$ rpm at time $t=0$ and $t=300$ s.
		(b) Position, $s$, of the same robot as a function of time, $t$. Solid line corresponds to the linear fit $s=v\,t$ where $v$ is the speed.
		(c) Position vs. time of a robot with $n=4$ and $\omega_{\textrm{total}}=208$ rpm, where stick-slip dynamics is prominent.
	}	
	\label{fig:Robotposition}
\end{figure*}

\section{Results and discussion}
\label{sec:Result}

Recall from Figure~\ref{fig:intro} that the motor embedded in the head generates a rotational speed, $\omega_T$. As a result, the head and the tails rotate along opposite directions with rotational speeds of $\omega_h$ and $\omega_t$ such that $\omega_T = \omega_h + \omega_t$ ($\omega_T, \omega_h,$ and $\omega_t$ are all non-negative values). The total rotational speed, $\omega_T$, is considered as a control parameter in our study. The resulting rotation of the tails generate a propulsive force along the axial direction (see Figure~\ref{fig:simSchematic}). This propulsive force is used by the head and the tail to overcome the drag from the granular medium and the entire system moves forward with a speed $v$.


\subsection{Threshold angular speed to move}
\label{sec:threshold}

When experiments were performed, it is found that there exists a threshold under which the robot stays still ($v=0$) and above which it starts to move. This can be understood from the mechanics of granular materials. Such materials can behave as a solid but also flow as a fluid. A threshold angular speed is necessary to transform the medium from solid-like behavior to fluid-like flow.
%
This threshold in our experiments is approximately $\omega_T \approx 50$ rpm  and thus there are no data points at $\omega_T \lesssim 50$ rpm in the forthcoming discussion. In this regime of solid-like behavior, the tails and the head still rotate relative to one another; however, the whole robot does not change its location.

In this study, the maximum total angular speed is $\omega_T \approx 250$ rpm and the aforementioned regime ($\omega_T \approx 50$) is a relatively small part of the the overall parameter space. In the simulations, we use RFT that does not consider this threshold. Nonetheless, the simulation can capture the motion of the robot when $v > 0$. It is relatively straightforward to include this threshold in Algorithm~\ref{algo:discreteElasticRods} by using conditional statements to impose boundary conditions on the head. However, this will introduce new fitting parameters without much improvement in the overall predictive ability of the simulation.

%

\subsection{Speed of the robot}
\label{experimentdata}

We use the speed of the robot, $v$, along the axial direction as the primary performance metric of the robot. This parameter will be used in subsequent sections to study the effect of the total angular speed, $\omega_T$, and the number of tails, $n$. The efficiency of the robot, $\eta$, will also be defined related with the speed, $v$.

During experiments, digital camera was used to capture videos of the motion of the robot. Figures~\ref{fig:Robotposition}(a1) and (a2) show two snapshots of a robot with $n=2$ tails and total rotational speed $\omega_T=250$ rpm at $t=0$ and $t=300$ sec. The green tails were marked with black markers and the black head was marked with bright yellow marker. Aided by the transparency of the granular medium and the markers on the robot, these videos were processed to extract the position of the robot, $s$, as a function of time. 
Figure~\ref{fig:Robotposition}(b) presents the position of the robot as a function of time. Closed triangles denote data from experiments and solid line represents a linear fit of the form $s=vt$. We observe that the robot moves at an almost constant velocity of $v \sim 0.6$ mm/s. This is expected from a solid body moving inside a medium governed by RFT.

On the other hand, Figure~\ref{fig:Robotposition}(c) shows the position of a robot with $n=4$ tails and rotational speed $\omega_T=208$ rpm. The motion of the robot is now qualitatively different from the one presented in Figure~\ref{fig:Robotposition}(b). The robot continuously moves forward in general but intermittently stays at the same position. This phenomenon is reminiscent of stick-slip -- sudden motion that occurs when two multiple bodies are sliding past one another. At larger number of tails (e.g. $n=4$ and $n=5$), experimental observations indicate that the granular medium can get jammed (i.e. increase in viscosity) and the robot frequently gets stuck. Interestingly, our experiments (see Figure~\ref{fig:Robotposition}b) indicate the robot can resolve the jamming on its own through rotation (i.e. creating disturbance) for a few seconds. The periodic jamming and stick-slip cannot be captured by RFT and we do not include this behavior in our simulations. We focus only on robots with $n=2$ and $n=3$ tails that move at a constant speed with time. Nonetheless, this indicates room for expanding the theories for locomotion inside granular medium beyond RFT. Integrating such theories that describe the viscosity as a function of the robot configuration and time into Algorithm~\ref{algo:discreteElasticRods} should be a relatively trivial task.

\begin{figure*}[t]
	\centering
	\includegraphics[width=0.9\linewidth]{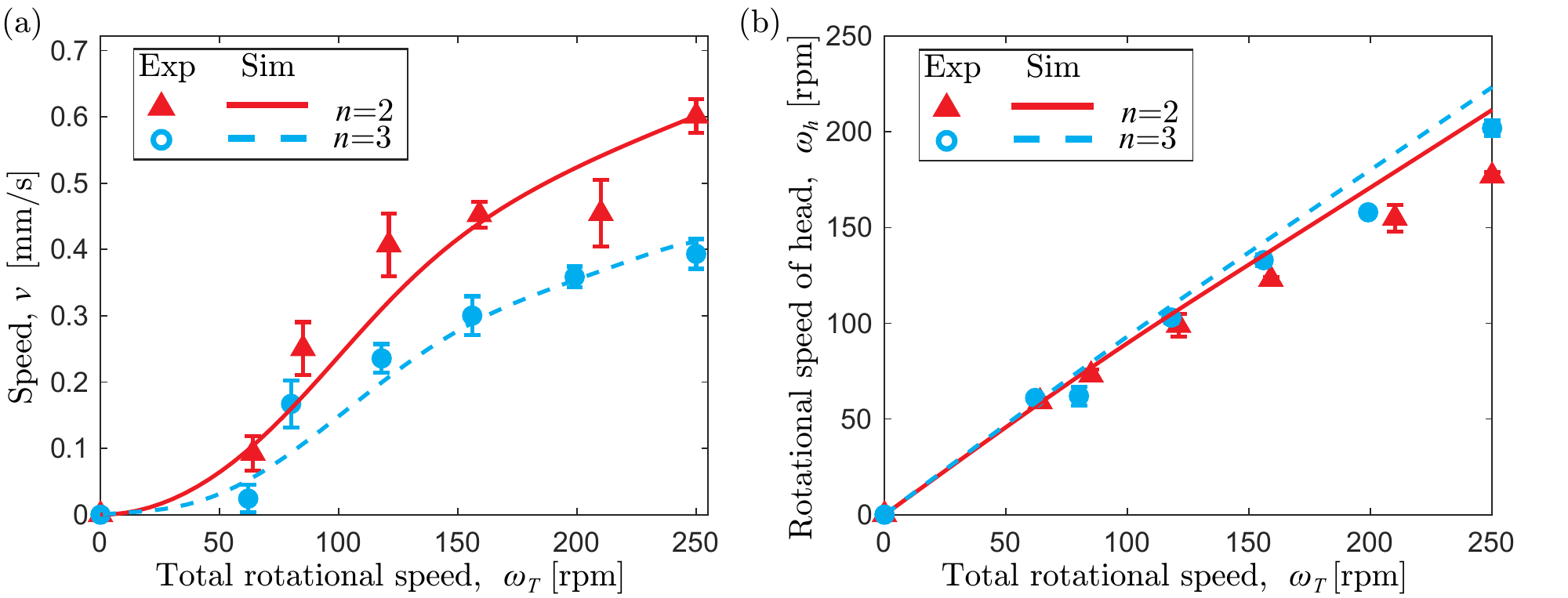}
	\caption{Comparison between experiment data and simulation results for the relationship between(a) total rotation speed of head and tail and robot moving speed; (b)total rotation speed of head and tail and rotation speed of head. The red triangles and blue circles with error bars are experiment data when the tail number is 2 and 3 respectively. The solid red line is the simulated outcome associated with the fitting parameters, $C_h$ and $\mu$ whereas the dashed blue line represents the simulation result predicted by the same fitting parameters.
	}
	\label{fig:velocity}
\end{figure*}

\subsection{Parameters fitting for simulations}
\label{subsection:parameterfitting}

We now move onto numerical simulations (details in Section~\ref{sec:modelingProcess}) to model the locomotion and deformation of the robot. Recall from Eqs. \ref{eq:stokes_drag_refined} and \ref{eq:stokes_torque_refined} that $C_1$ and $C_2$ are fitting parameters to account for the shape and surface roughness of the robot head. In addition, $\mu$ is the third fitting parameter standing for the dynamic viscosity of the granular medium. As detailed next, experimental data with a 2-tailed robot ($n=2$) are used to estimate $C_1, C_2,$ and $\mu$. Simulations are performed with these parameters for the $n=3$ case; simulation results are then compared against experiments for validation of the fitting process.

Figures~\ref{fig:velocity}(a) and (b) present the speed of the robot, $v$, and the rotational speed of the head, $\omega_h$, respectively, as functions of the total rotational speed, $\omega_T$. The data for $n=2$ and $n=3$ are shown in the figures. The data ($v$ vs. $\omega_T$ and $\omega_h$ vs. $\omega_T$) for $n=2$ are used to obtain the best fit values of the fitting parameters: $C_1=2.420, C_2=0.039$, and $\mu=6.828$.
These parameters are then used in the numerical tool to simulate the locomotion of a robot with three tails. In Figure~\ref{fig:velocity}(a), speed vs. total rotational speed data show good agreement between experiments and simulations. Figure~\ref{fig:velocity}(b) shows the rotational speed of the head as a function of total rotational speed and we find that, in both experiments and simulations, a robot with $n=3$ has a slightly larger head rotational speed than the one with $n=2$.


The slight mismatch between the experimental results and simulation data can be partially attributed to the various assumptions made in the model. The fluid model assumes that the drag force exerted by the granular medium can be expressed using RFT. The structure model assumes that the tails are infinitesimally thin elastic rods. The drag force on the head is assumed to be linearly proportional to the velocity and the torque on the head is linearly proportional to its angular speed. In addition, invariably there are experimental errors, e.g. structural defects introduced during fabrication. Nonetheless, the reasonably good agreement between experiments and simulations support the validity of RFT in this case.

\subsection{Speed vs. number of tails}
\label{sec:speedVtails}

A counter intuitive observation from Figure~\ref{fig:velocity}(a) is that, at a fixed value of rotational speed of the motor, $\omega_T$, the robot with 2 tails moves faster than the one with 3 tails. Additionally, the speed vs. total rotational speed curve is nonlinear. All of these point to the large structural deformation and the strong coupling between the head and the tails.

For a physical understanding of the dependence of speed on the number of tails, let us first note the rotational speed of the head for the two cases in Figure~\ref{fig:velocity}(b). As the number of tails, $n$, increases, the rotational speed of the head, $\omega_h$, increases (at a fixed value of $\omega_T$). Since $\omega_T = \omega_h + \omega_t$, this implies that the rotational speed of the tails, $\omega_t$, decreases as the number of tails, $n$, increases. The propulsive force generated by each tail -- let us denote this quantity as $f_t$ -- therefore also decreases. However, two additional factors to be considered to understand the overall speed, $v$, of the robot. First, the total propulsive force available is $n \, f_t$ and even though increasing $n$ reduces $f_t$, it may (or may not) ultimately increase $n \, f_t$. Second, the total propulsive force is spent to overcome the drag on the head and the tail. As $n$ increases, the amount of propulsive force spent on moving the tails forward also increases and the propulsive force budgeted for the head decreases. All of the these factors above combined dictate the dependence between the robot speed and the number of tails. 

In the experiments presented herein, the set of physical parameters are such that the speed decreases with the number of tails. However, this is not universally true for this system. For example, consider a robot with $C_2 \to \infty$ such that the head never rotates (i.e. $\omega_h = 0$). In that case, the rotational speed of the tail is always equal to the total rotation speed and $f_t$ is a function of only $\omega_T$ (and not $n$). Then, the total propulsive force, $n\, f_t$, increases with $n$ (assuming $\omega_T$ is fixed) and the speed of the robot is likely going to increase.

\begin{figure}[h]
	\centering
	\includegraphics[width=0.9\linewidth]{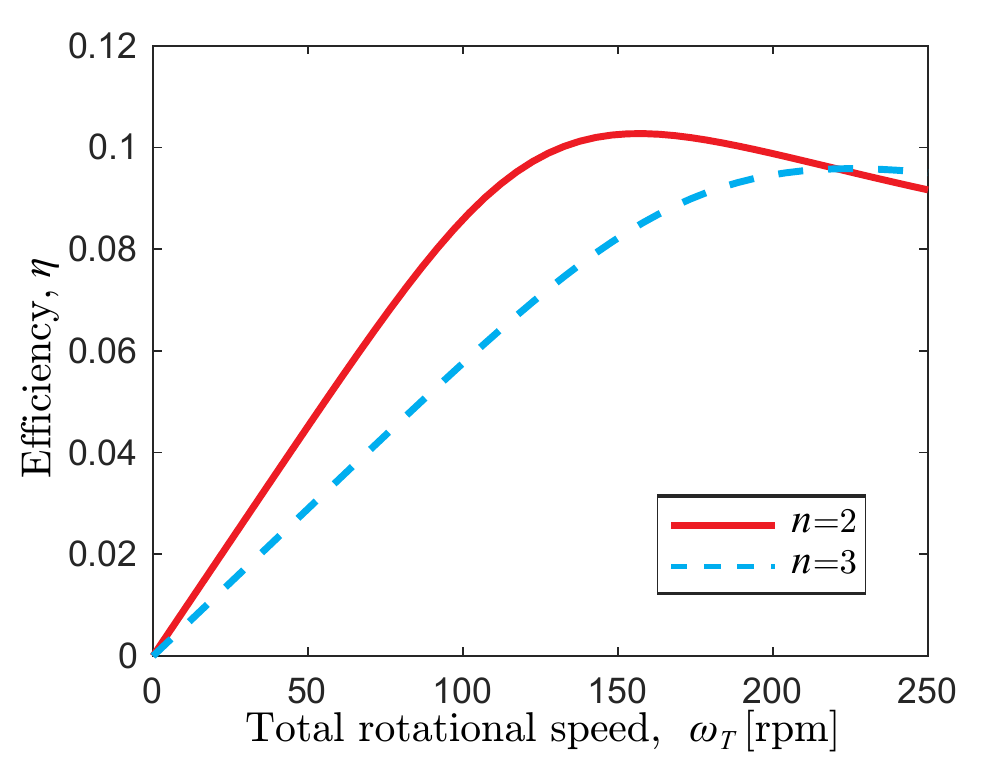}
	\caption{Variation of torque utilization efficiency versus the total rotational speed of robot predicted by our simulator.}
	\label{fig:efficiency}
\end{figure}

\subsection{Efficiency}
\label{sec:efficiency}
The efficiency, $\eta$, of the robot is defined as the ratio of propulsive force to propulsive torque. Since $\eta$ is a non-dimensional quantity, we choose the radius of the head, $a$, as the length-scale. The expression for $\eta$ is
\begin{equation}
\eta 
= \frac{ |\mathbf F_h|}{ |\mathbf T_h| }\, a
= \frac {6\pi C_1 \mu a^2 v} {8 \pi C_2  \mu a^3 \omega_h},
\label{eq:efficiency}
\end{equation}
where $| \cdot | $ denotes absolute value and expressions for $\mathbf F_h$ and $\mathbf T_h$ can be found in Eqs.~\ref{eq:stokes_drag_refined} and ~\ref{eq:stokes_torque_refined}, respectively.
The numerator can be understood as the drag force exerted by the medium on the robot while the denominator gives the total torque  generated by the rotation of the motor. We use the simulator in Figure \ref{fig:efficiency} to predict the variation of efficiency, $\eta$, with the rotational speed, $\omega_T$, of the robot. The efficiency of a robot with $n=2$ is non-monotonic and reaches the maximum at $\omega_T \approx 150$ rpm. At this optimal rotational speed, the robot moves the farthest per unit torque from the motor. Such clear presence of an optimal rotational speed in the operating range of the motor highlights the need for a numerical simulator that can be used as a design tool for the robot. Moreover, for $\omega_T \lesssim 200$rpm, the efficiency of a 2-tailed robot is greater than the one of a 3-tailed robot. Beyond  $\omega_T \gtrsim 200$rpm, the 3-tailed one performs better than the robot with 2 tails. These observations underline the high degree of nonlinearity in the functional dependence between the efficiency and the physical parameters (e.g. $n$ and $\omega_T$).

\section{Conclusion}
\label{sec:Conclusion}

In this work, a discrete differential geometry-based simulation framework was introduced that models the robot as a composition of Kirchhoff elastic rod. The robot is discretized into a system of mass-spring system, with discrete elastic (bending, twisting, stretching) energies associated with each spring. The total elastic energy of the robot is the sum of all the discrete elastic energies. Equations of motion are formulated that are simply statements of the following: at each DOF, the sum of elastic force (i.e. negative gradient of the elastic energy) and external force is equal to the lumped mass times acceleration of that DOF. The actuation of the robot (i.e. rotational speed of the motor) is represented by a time varying natural strain (specifically, the natural twist at the node representing the head). This approach allows us to simulate the shape of the robot in a fully implicit (Euler backward) manner.

The external force in this setup is the drag force exerted by the granular medium on the robotic structure. RFT -- originally developed to model the hydrodynamics of low Reynolds fluid flow -- was used to model the external force by the granular medium. This force can be seamless integrated into the simulation framework. Comparison between experiments and simulations showed that RFT is reasonably valid in case of the flagellated robot discussed here. However, when the number of tails is large (and the spacing between tails is small), ``stick-slip" was observed and the underlying assumption behind RFT was no longer valid. In the future, the drag model of the granular medium can be improved to account for such behavior.

The simulation tool, supported by experiments, shed light on the highly nonlinear functional dependence between the performance of the robot (e.g. speed or efficiency) and the relevant physical parameters (e.g. the number of tails). Some counter-intuitive observations include the inverse relation between the speed and the number of tails of the robot in the representative setup. The non-monotonic dependence of efficiency on the rotational speed of the motor highlighted the necessity of a design tool for optimal control of the robot. The computational speed of the simulator can be exploited to run parametric studies and identify the optimal design and control of this class of robots that can be represented by stick figures.

\section{Funding}
This work was supported by the Henry Samueli School of Engineering and Applied Science, University of California and the National Science Foundation (Award \# IIS-1925360).


\begin{thebibliography}{52}
\expandafter\ifx\csname natexlab\endcsname\relax\def\natexlab#1{#1}\fi
\providecommand{\url}[1]{\texttt{#1}}
\providecommand{\href}[2]{#2}
\providecommand{\path}[1]{#1}
\providecommand{\DOIprefix}{doi:}
\providecommand{\ArXivprefix}{arXiv:}
\providecommand{\URLprefix}{URL: }
\providecommand{\Pubmedprefix}{pmid:}
\providecommand{\doi}[1]{\href{http://dx.doi.org/#1}{\path{#1}}}
\providecommand{\Pubmed}[1]{\href{pmid:#1}{\path{#1}}}
\providecommand{\bibinfo}[2]{#2}
\ifx\xfnm\relax \def\xfnm[#1]{\unskip,\space#1}\fi
\bibitem[{Alexander(2003)}]{alexander2003principles}
\bibinfo{author}{Alexander, R.M.}, \bibinfo{year}{2003}.
\newblock \bibinfo{title}{Principles of animal locomotion}.
\newblock \bibinfo{publisher}{Princeton University Press}.
\bibitem[{Audoly and Pomeau(2000)}]{audoly2000elasticity}
\bibinfo{author}{Audoly, B.}, \bibinfo{author}{Pomeau, Y.},
  \bibinfo{year}{2000}.
\newblock \bibinfo{title}{Elasticity and geometry}, in:
  \bibinfo{booktitle}{Peyresq Lectures on Nonlinear Phenomena}.
  \bibinfo{publisher}{World Scientific}, pp. \bibinfo{pages}{1--35}.
\bibitem[{Baek et~al.(2018)Baek, Sageman-Furnas, Jawed and Reis}]{baek2018form}
\bibinfo{author}{Baek, C.}, \bibinfo{author}{Sageman-Furnas, A.O.},
  \bibinfo{author}{Jawed, M.K.}, \bibinfo{author}{Reis, P.M.},
  \bibinfo{year}{2018}.
\newblock \bibinfo{title}{Form finding in elastic gridshells}.
\newblock \bibinfo{journal}{Proceedings of the National Academy of Sciences}
  \bibinfo{volume}{115}, \bibinfo{pages}{75--80}.
\bibitem[{Bartlett et~al.(2017)Bartlett, Kazem, Powell-Palm, Huang, Sun, Malen
  and Majidi}]{bartlett2017high}
\bibinfo{author}{Bartlett, M.D.}, \bibinfo{author}{Kazem, N.},
  \bibinfo{author}{Powell-Palm, M.J.}, \bibinfo{author}{Huang, X.},
  \bibinfo{author}{Sun, W.}, \bibinfo{author}{Malen, J.A.},
  \bibinfo{author}{Majidi, C.}, \bibinfo{year}{2017}.
\newblock \bibinfo{title}{High thermal conductivity in soft elastomers with
  elongated liquid metal inclusions}.
\newblock \bibinfo{journal}{Proceedings of the National Academy of Sciences} ,
  \bibinfo{pages}{201616377}.
\bibitem[{Bergou et~al.(2010)Bergou, Audoly, Vouga, Wardetzky and
  Grinspun}]{bergou2010discrete}
\bibinfo{author}{Bergou, M.}, \bibinfo{author}{Audoly, B.},
  \bibinfo{author}{Vouga, E.}, \bibinfo{author}{Wardetzky, M.},
  \bibinfo{author}{Grinspun, E.}, \bibinfo{year}{2010}.
\newblock \bibinfo{title}{Discrete viscous threads}, in:
  \bibinfo{booktitle}{ACM Transactions on Graphics},
  \bibinfo{organization}{ACM}. p. \bibinfo{pages}{116}.
\bibitem[{Bergou et~al.(2008)Bergou, Wardetzky, Robinson, Audoly and
  Grinspun}]{bergou2008discrete}
\bibinfo{author}{Bergou, M.}, \bibinfo{author}{Wardetzky, M.},
  \bibinfo{author}{Robinson, S.}, \bibinfo{author}{Audoly, B.},
  \bibinfo{author}{Grinspun, E.}, \bibinfo{year}{2008}.
\newblock \bibinfo{title}{Discrete elastic rods}.
\newblock \bibinfo{journal}{ACM transactions on graphics (TOG)}
  \bibinfo{volume}{27}, \bibinfo{pages}{63}.
\bibitem[{Biewener(1990)}]{biewener1990biomechanics}
\bibinfo{author}{Biewener, A.A.}, \bibinfo{year}{1990}.
\newblock \bibinfo{title}{Biomechanics of mammalian terrestrial locomotion}.
\newblock \bibinfo{journal}{Science} \bibinfo{volume}{250},
  \bibinfo{pages}{1097--1103}.
\bibitem[{Choi et~al.(2021)Choi, Tong, Jawed and Joo}]{choi2021implicit}
\bibinfo{author}{Choi, A.}, \bibinfo{author}{Tong, D.}, \bibinfo{author}{Jawed,
  M.K.}, \bibinfo{author}{Joo, J.}, \bibinfo{year}{2021}.
\newblock \bibinfo{title}{Implicit contact model for discrete elastic rods in
  knot tying}.
\newblock \bibinfo{journal}{Journal of Applied Mechanics} ,
  \bibinfo{pages}{1--13}.
\bibitem[{Conte et~al.(2010)Conte, Modarres-Sadeghi, Watts, Hover and
  Triantafyllou}]{conte2010fast}
\bibinfo{author}{Conte, J.}, \bibinfo{author}{Modarres-Sadeghi, Y.},
  \bibinfo{author}{Watts, M.}, \bibinfo{author}{Hover, F.S.},
  \bibinfo{author}{Triantafyllou, M.S.}, \bibinfo{year}{2010}.
\newblock \bibinfo{title}{A fast-starting mechanical fish that accelerates at
  40 ms- 2}.
\newblock \bibinfo{journal}{Bioinspiration \& biomimetics} \bibinfo{volume}{5},
  \bibinfo{pages}{035004}.
\bibitem[{Dickinson et~al.(2000)Dickinson, Farley, Full, Koehl, Kram and
  Lehman}]{dickinson2000animals}
\bibinfo{author}{Dickinson, M.H.}, \bibinfo{author}{Farley, C.T.},
  \bibinfo{author}{Full, R.J.}, \bibinfo{author}{Koehl, M.},
  \bibinfo{author}{Kram, R.}, \bibinfo{author}{Lehman, S.},
  \bibinfo{year}{2000}.
\newblock \bibinfo{title}{How animals move: an integrative view}.
\newblock \bibinfo{journal}{science} \bibinfo{volume}{288},
  \bibinfo{pages}{100--106}.
\bibitem[{Ding et~al.(2012)Ding, Sharpe, Masse and Goldman}]{ding2012mechanics}
\bibinfo{author}{Ding, Y.}, \bibinfo{author}{Sharpe, S.S.},
  \bibinfo{author}{Masse, A.}, \bibinfo{author}{Goldman, D.I.},
  \bibinfo{year}{2012}.
\newblock \bibinfo{title}{Mechanics of undulatory swimming in a frictional
  fluid}.
\newblock \bibinfo{journal}{PLoS computational biology} \bibinfo{volume}{8},
  \bibinfo{pages}{e1002810}.
\bibitem[{Forghani et~al.(2021)Forghani, Huang and Jawed}]{forghani2021control}
\bibinfo{author}{Forghani, M.}, \bibinfo{author}{Huang, W.},
  \bibinfo{author}{Jawed, M.K.}, \bibinfo{year}{2021}.
\newblock \bibinfo{title}{Control of uniflagellar soft robots at low reynolds
  number using buckling instability}.
\newblock \bibinfo{journal}{Journal of Dynamic Systems, Measurement, and
  Control} \bibinfo{volume}{143}, \bibinfo{pages}{061004}.
\bibitem[{Gray(1968)}]{gray1968animal}
\bibinfo{author}{Gray, J.}, \bibinfo{year}{1968}.
\newblock \bibinfo{title}{Animal locomotion}.
\newblock \bibinfo{publisher}{Weidenfeld \& Nicolson}.
\bibitem[{Gray and Hancock(1955)}]{gray1955propulsion}
\bibinfo{author}{Gray, J.}, \bibinfo{author}{Hancock, G.},
  \bibinfo{year}{1955}.
\newblock \bibinfo{title}{The propulsion of sea-urchin spermatozoa}.
\newblock \bibinfo{journal}{Journal of Experimental Biology}
  \bibinfo{volume}{32}, \bibinfo{pages}{802--814}.
\bibitem[{Hu et~al.(2009)Hu, Nirody, Scott and Shelley}]{hu2009mechanics}
\bibinfo{author}{Hu, D.L.}, \bibinfo{author}{Nirody, J.},
  \bibinfo{author}{Scott, T.}, \bibinfo{author}{Shelley, M.J.},
  \bibinfo{year}{2009}.
\newblock \bibinfo{title}{The mechanics of slithering locomotion}.
\newblock \bibinfo{journal}{Proceedings of the National Academy of Sciences}
  \bibinfo{volume}{106}, \bibinfo{pages}{10081--10085}.
\bibitem[{Jawed and Reis(2017)}]{jawed2017dynamics}
\bibinfo{author}{Jawed, M.}, \bibinfo{author}{Reis, P.M.},
  \bibinfo{year}{2017}.
\newblock \bibinfo{title}{Dynamics of a flexible helical filament rotating in a
  viscous fluid near a rigid boundary}.
\newblock \bibinfo{journal}{Physical Review Fluids} \bibinfo{volume}{2},
  \bibinfo{pages}{034101}.
\bibitem[{Jawed et~al.(2014)Jawed, Da, Joo, Grinspun and
  Reis}]{jawed2014coiling}
\bibinfo{author}{Jawed, M.K.}, \bibinfo{author}{Da, F.}, \bibinfo{author}{Joo,
  J.}, \bibinfo{author}{Grinspun, E.}, \bibinfo{author}{Reis, P.M.},
  \bibinfo{year}{2014}.
\newblock \bibinfo{title}{Coiling of elastic rods on rigid substrates}.
\newblock \bibinfo{journal}{Proceedings of the National Academy of Sciences}
  \bibinfo{volume}{111}, \bibinfo{pages}{14663--14668}.
\bibitem[{Jawed et~al.(2015)Jawed, Khouri, Da, Grinspun and
  Reis}]{jawed2015propulsion}
\bibinfo{author}{Jawed, M.K.}, \bibinfo{author}{Khouri, N.},
  \bibinfo{author}{Da, F.}, \bibinfo{author}{Grinspun, E.},
  \bibinfo{author}{Reis, P.M.}, \bibinfo{year}{2015}.
\newblock \bibinfo{title}{Propulsion and instability of a flexible helical rod
  rotating in a viscous fluid}.
\newblock \bibinfo{journal}{Physical review letters} \bibinfo{volume}{115},
  \bibinfo{pages}{168101}.
\bibitem[{Jawed et~al.(2018)Jawed, Novelia and O'Reilly}]{jawed2018primer}
\bibinfo{author}{Jawed, M.K.}, \bibinfo{author}{Novelia, A.},
  \bibinfo{author}{O'Reilly, O.M.}, \bibinfo{year}{2018}.
\newblock \bibinfo{title}{A Primer on the Kinematics of Discrete Elastic Rods}.
\newblock \bibinfo{publisher}{Springer}.
\bibitem[{Jawed and Reis(2016)}]{jawed2016deformation}
\bibinfo{author}{Jawed, M.K.}, \bibinfo{author}{Reis, P.M.},
  \bibinfo{year}{2016}.
\newblock \bibinfo{title}{Deformation of a soft helical filament in an axial
  flow at low reynolds number}.
\newblock \bibinfo{journal}{Soft matter} \bibinfo{volume}{12},
  \bibinfo{pages}{1898--1905}.
\bibitem[{Johnson and Brokaw(1979)}]{johnson1979flagellar}
\bibinfo{author}{Johnson, R.}, \bibinfo{author}{Brokaw, C.},
  \bibinfo{year}{1979}.
\newblock \bibinfo{title}{Flagellar hydrodynamics. a comparison between
  resistive-force theory and slender-body theory}.
\newblock \bibinfo{journal}{Biophysical journal} \bibinfo{volume}{25},
  \bibinfo{pages}{113--127}.
\bibitem[{Kim et~al.(2013)Kim, Laschi and Trimmer}]{kim2013soft}
\bibinfo{author}{Kim, S.}, \bibinfo{author}{Laschi, C.},
  \bibinfo{author}{Trimmer, B.}, \bibinfo{year}{2013}.
\newblock \bibinfo{title}{Soft robotics: a bioinspired evolution in robotics}.
\newblock \bibinfo{journal}{Trends in biotechnology} \bibinfo{volume}{31},
  \bibinfo{pages}{287--294}.
\bibitem[{Kirchhoff(1859)}]{kirchhoff1859uber}
\bibinfo{author}{Kirchhoff, G.}, \bibinfo{year}{1859}.
\newblock \bibinfo{title}{Uber das gleichgewicht und die bewegung eines
  unendlich dunnen elastischen stabes}.
\newblock \bibinfo{journal}{J. Reine Angew. Math.} \bibinfo{volume}{56},
  \bibinfo{pages}{285--313}.
\bibitem[{Laschi et~al.(2012)Laschi, Cianchetti, Mazzolai, Margheri, Follador
  and Dario}]{laschi2012soft}
\bibinfo{author}{Laschi, C.}, \bibinfo{author}{Cianchetti, M.},
  \bibinfo{author}{Mazzolai, B.}, \bibinfo{author}{Margheri, L.},
  \bibinfo{author}{Follador, M.}, \bibinfo{author}{Dario, P.},
  \bibinfo{year}{2012}.
\newblock \bibinfo{title}{Soft robot arm inspired by the octopus}.
\newblock \bibinfo{journal}{Advanced Robotics} \bibinfo{volume}{26},
  \bibinfo{pages}{709--727}.
\bibitem[{Lauga(2011)}]{lauga2011life}
\bibinfo{author}{Lauga, E.}, \bibinfo{year}{2011}.
\newblock \bibinfo{title}{Life around the scallop theorem}.
\newblock \bibinfo{journal}{Soft Matter} \bibinfo{volume}{7},
  \bibinfo{pages}{3060--3065}.
\bibitem[{Lauga and Powers(2009)}]{lauga2009hydrodynamics}
\bibinfo{author}{Lauga, E.}, \bibinfo{author}{Powers, T.R.},
  \bibinfo{year}{2009}.
\newblock \bibinfo{title}{The hydrodynamics of swimming microorganisms}.
\newblock \bibinfo{journal}{Reports on Progress in Physics}
  \bibinfo{volume}{72}, \bibinfo{pages}{096601}.
\bibitem[{Lazarus et~al.(2013)Lazarus, Miller, Metlitz and
  Reis}]{lazarus2013contorting}
\bibinfo{author}{Lazarus, A.}, \bibinfo{author}{Miller, J.T.},
  \bibinfo{author}{Metlitz, M.M.}, \bibinfo{author}{Reis, P.M.},
  \bibinfo{year}{2013}.
\newblock \bibinfo{title}{Contorting a heavy and naturally curved elastic rod}.
\newblock \bibinfo{journal}{Soft Matter} \bibinfo{volume}{9},
  \bibinfo{pages}{8274--8281}.
\bibitem[{Licht et~al.(2004)Licht, Polidoro, Flores, Hover and
  Triantafyllou}]{licht2004design}
\bibinfo{author}{Licht, S.}, \bibinfo{author}{Polidoro, V.},
  \bibinfo{author}{Flores, M.}, \bibinfo{author}{Hover, F.S.},
  \bibinfo{author}{Triantafyllou, M.S.}, \bibinfo{year}{2004}.
\newblock \bibinfo{title}{Design and projected performance of a flapping foil
  auv}.
\newblock \bibinfo{journal}{IEEE Journal of oceanic engineering}
  \bibinfo{volume}{29}, \bibinfo{pages}{786--794}.
\bibitem[{Lighthill(1976)}]{lighthill1976flagellar}
\bibinfo{author}{Lighthill, J.}, \bibinfo{year}{1976}.
\newblock \bibinfo{title}{Flagellar hydrodynamics}.
\newblock \bibinfo{journal}{SIAM review} \bibinfo{volume}{18},
  \bibinfo{pages}{161--230}.
\bibitem[{Lin et~al.(2011)Lin, Leisk and Trimmer}]{lin2011goqbot}
\bibinfo{author}{Lin, H.T.}, \bibinfo{author}{Leisk, G.G.},
  \bibinfo{author}{Trimmer, B.}, \bibinfo{year}{2011}.
\newblock \bibinfo{title}{Goqbot: a caterpillar-inspired soft-bodied rolling
  robot}.
\newblock \bibinfo{journal}{Bioinspiration \& biomimetics} \bibinfo{volume}{6},
  \bibinfo{pages}{026007}.
\bibitem[{Macnab and Ornston(1977)}]{macnab1977normal}
\bibinfo{author}{Macnab, R.M.}, \bibinfo{author}{Ornston, M.K.},
  \bibinfo{year}{1977}.
\newblock \bibinfo{title}{Normal-to-curly flagellar transitions and their role
  in bacterial tumbling. stabilization of an alternative quaternary structure
  by mechanical force}.
\newblock \bibinfo{journal}{Journal of molecular biology}
  \bibinfo{volume}{112}, \bibinfo{pages}{1--30}.
\bibitem[{Majidi et~al.(2013)Majidi, Shepherd, Kramer, Whitesides and
  Wood}]{majidi2013influence}
\bibinfo{author}{Majidi, C.}, \bibinfo{author}{Shepherd, R.F.},
  \bibinfo{author}{Kramer, R.K.}, \bibinfo{author}{Whitesides, G.M.},
  \bibinfo{author}{Wood, R.J.}, \bibinfo{year}{2013}.
\newblock \bibinfo{title}{Influence of surface traction on soft robot
  undulation}.
\newblock \bibinfo{journal}{The International Journal of Robotics Research}
  \bibinfo{volume}{32}, \bibinfo{pages}{1577--1584}.
\bibitem[{Maladen et~al.(2009)Maladen, Ding, Li and
  Goldman}]{maladen2009undulatory}
\bibinfo{author}{Maladen, R.D.}, \bibinfo{author}{Ding, Y.},
  \bibinfo{author}{Li, C.}, \bibinfo{author}{Goldman, D.I.},
  \bibinfo{year}{2009}.
\newblock \bibinfo{title}{Undulatory swimming in sand: subsurface locomotion of
  the sandfish lizard}.
\newblock \bibinfo{journal}{science} \bibinfo{volume}{325},
  \bibinfo{pages}{314--318}.
\bibitem[{Maladen et~al.(2011a)Maladen, Ding, Umbanhowar and
  Goldman}]{maladen2011undulatory}
\bibinfo{author}{Maladen, R.D.}, \bibinfo{author}{Ding, Y.},
  \bibinfo{author}{Umbanhowar, P.B.}, \bibinfo{author}{Goldman, D.I.},
  \bibinfo{year}{2011}a.
\newblock \bibinfo{title}{Undulatory swimming in sand: experimental and
  simulation studies of a robotic sandfish}.
\newblock \bibinfo{journal}{The International Journal of Robotics Research}
  \bibinfo{volume}{30}, \bibinfo{pages}{793--805}.
\bibitem[{Maladen et~al.(2011b)Maladen, Ding, Umbanhowar, Kamor and
  Goldman}]{maladen2011mechanical}
\bibinfo{author}{Maladen, R.D.}, \bibinfo{author}{Ding, Y.},
  \bibinfo{author}{Umbanhowar, P.B.}, \bibinfo{author}{Kamor, A.},
  \bibinfo{author}{Goldman, D.I.}, \bibinfo{year}{2011}b.
\newblock \bibinfo{title}{Mechanical models of sandfish locomotion reveal
  principles of high performance subsurface sand-swimming}.
\newblock \bibinfo{journal}{Journal of The Royal Society Interface}
  \bibinfo{volume}{8}, \bibinfo{pages}{1332--1345}.
\bibitem[{Miller et~al.(2014)Miller, Lazarus, Audoly and
  Reis}]{miller2014shapes}
\bibinfo{author}{Miller, J.}, \bibinfo{author}{Lazarus, A.},
  \bibinfo{author}{Audoly, B.}, \bibinfo{author}{Reis, P.M.},
  \bibinfo{year}{2014}.
\newblock \bibinfo{title}{Shapes of a suspended curly hair}.
\newblock \bibinfo{journal}{Physical review letters} \bibinfo{volume}{112},
  \bibinfo{pages}{068103}.
\bibitem[{Panetta et~al.(2019)Panetta, Konakovi{\'c}-Lukovi{\'c}, Isvoranu,
  Bouleau and Pauly}]{panetta2019x}
\bibinfo{author}{Panetta, J.}, \bibinfo{author}{Konakovi{\'c}-Lukovi{\'c}, M.},
  \bibinfo{author}{Isvoranu, F.}, \bibinfo{author}{Bouleau, E.},
  \bibinfo{author}{Pauly, M.}, \bibinfo{year}{2019}.
\newblock \bibinfo{title}{X-shells: A new class of deployable beam structures}.
\newblock \bibinfo{journal}{ACM Transactions on Graphics (TOG)}
  \bibinfo{volume}{38}, \bibinfo{pages}{1--15}.
\bibitem[{P{\'e}rez et~al.(2015)P{\'e}rez, Thomaszewski, Coros, Bickel,
  Canabal, Sumner and Otaduy}]{perez2015design}
\bibinfo{author}{P{\'e}rez, J.}, \bibinfo{author}{Thomaszewski, B.},
  \bibinfo{author}{Coros, S.}, \bibinfo{author}{Bickel, B.},
  \bibinfo{author}{Canabal, J.A.}, \bibinfo{author}{Sumner, R.},
  \bibinfo{author}{Otaduy, M.A.}, \bibinfo{year}{2015}.
\newblock \bibinfo{title}{Design and fabrication of flexible rod meshes}.
\newblock \bibinfo{journal}{ACM Transactions on Graphics} \bibinfo{volume}{34},
  \bibinfo{pages}{138}.
\bibitem[{Renda et~al.(2018)Renda, Giorgio-Serchi, Boyer, Laschi, Dias and
  Seneviratne}]{renda2018unified}
\bibinfo{author}{Renda, F.}, \bibinfo{author}{Giorgio-Serchi, F.},
  \bibinfo{author}{Boyer, F.}, \bibinfo{author}{Laschi, C.},
  \bibinfo{author}{Dias, J.}, \bibinfo{author}{Seneviratne, L.},
  \bibinfo{year}{2018}.
\newblock \bibinfo{title}{A unified multi-soft-body dynamic model for
  underwater soft robots}.
\newblock \bibinfo{journal}{The International Journal of Robotics Research}
  \bibinfo{volume}{37}, \bibinfo{pages}{648--666}.
\bibitem[{Rodenborn et~al.(2013)Rodenborn, Chen, Swinney, Liu and
  Zhang}]{rodenborn2013propulsion}
\bibinfo{author}{Rodenborn, B.}, \bibinfo{author}{Chen, C.H.},
  \bibinfo{author}{Swinney, H.L.}, \bibinfo{author}{Liu, B.},
  \bibinfo{author}{Zhang, H.}, \bibinfo{year}{2013}.
\newblock \bibinfo{title}{Propulsion of microorganisms by a helical flagellum}.
\newblock \bibinfo{journal}{Proceedings of the National Academy of Sciences}
  \bibinfo{volume}{110}, \bibinfo{pages}{E338--E347}.
\bibitem[{Rus and Tolley(2015)}]{rus2015design}
\bibinfo{author}{Rus, D.}, \bibinfo{author}{Tolley, M.T.},
  \bibinfo{year}{2015}.
\newblock \bibinfo{title}{Design, fabrication and control of soft robots}.
\newblock \bibinfo{journal}{Nature} \bibinfo{volume}{521},
  \bibinfo{pages}{467}.
\bibitem[{Saimek and Li(2004)}]{saimek2004motion}
\bibinfo{author}{Saimek, S.}, \bibinfo{author}{Li, P.Y.}, \bibinfo{year}{2004}.
\newblock \bibinfo{title}{Motion planning and control of a swimming machine}.
\newblock \bibinfo{journal}{The International Journal of Robotics Research}
  \bibinfo{volume}{23}, \bibinfo{pages}{27--53}.
\bibitem[{Scaramuzza et~al.(2009)Scaramuzza, Siegwart and
  Martinelli}]{scaramuzza2009international}
\bibinfo{author}{Scaramuzza, D.}, \bibinfo{author}{Siegwart, R.},
  \bibinfo{author}{Martinelli, A.}, \bibinfo{year}{2009}.
\newblock \bibinfo{title}{The international journal of robotics research}.
\newblock \bibinfo{journal}{The International Journal of Robotics Research}
  \bibinfo{volume}{28}, \bibinfo{pages}{149--171}.
\bibitem[{Schenk and G{\"a}rtner(2002)}]{schenk2002solving}
\bibinfo{author}{Schenk, O.}, \bibinfo{author}{G{\"a}rtner, K.},
  \bibinfo{year}{2002}.
\newblock \bibinfo{title}{Solving unsymmetric sparse systems of linear
  equations with pardiso}, in: \bibinfo{booktitle}{International Conference on
  Computational Science}, \bibinfo{organization}{Springer}. pp.
  \bibinfo{pages}{355--363}.
\bibitem[{Shepherd et~al.(2011)Shepherd, Ilievski, Choi, Morin, Stokes, Mazzeo,
  Chen, Wang and Whitesides}]{shepherd2011multigait}
\bibinfo{author}{Shepherd, R.F.}, \bibinfo{author}{Ilievski, F.},
  \bibinfo{author}{Choi, W.}, \bibinfo{author}{Morin, S.A.},
  \bibinfo{author}{Stokes, A.A.}, \bibinfo{author}{Mazzeo, A.D.},
  \bibinfo{author}{Chen, X.}, \bibinfo{author}{Wang, M.},
  \bibinfo{author}{Whitesides, G.M.}, \bibinfo{year}{2011}.
\newblock \bibinfo{title}{Multigait soft robot}.
\newblock \bibinfo{journal}{Proceedings of the national academy of sciences}
  \bibinfo{volume}{108}, \bibinfo{pages}{20400--20403}.
\bibitem[{Son et~al.(2013)Son, Guasto and Stocker}]{son2013bacteria}
\bibinfo{author}{Son, K.}, \bibinfo{author}{Guasto, J.S.},
  \bibinfo{author}{Stocker, R.}, \bibinfo{year}{2013}.
\newblock \bibinfo{title}{Bacteria can exploit a flagellar buckling instability
  to change direction}.
\newblock \bibinfo{journal}{Nature physics} \bibinfo{volume}{9},
  \bibinfo{pages}{494}.
\bibitem[{Taylor et~al.(2003)Taylor, Nudds and Thomas}]{taylor2003flying}
\bibinfo{author}{Taylor, G.K.}, \bibinfo{author}{Nudds, R.L.},
  \bibinfo{author}{Thomas, A.L.}, \bibinfo{year}{2003}.
\newblock \bibinfo{title}{Flying and swimming animals cruise at a strouhal
  number tuned for high power efficiency}.
\newblock \bibinfo{journal}{Nature} \bibinfo{volume}{425},
  \bibinfo{pages}{707}.
\bibitem[{Texier et~al.(2017)Texier, Ibarra and Melo}]{texier2017helical}
\bibinfo{author}{Texier, B.D.}, \bibinfo{author}{Ibarra, A.},
  \bibinfo{author}{Melo, F.}, \bibinfo{year}{2017}.
\newblock \bibinfo{title}{Helical locomotion in a granular medium}.
\newblock \bibinfo{journal}{Physical review letters} \bibinfo{volume}{119},
  \bibinfo{pages}{068003}.
\bibitem[{Thawani and Tirumkudulu(2018)}]{thawani2018trajectory}
\bibinfo{author}{Thawani, A.}, \bibinfo{author}{Tirumkudulu, M.S.},
  \bibinfo{year}{2018}.
\newblock \bibinfo{title}{Trajectory of a model bacterium}.
\newblock \bibinfo{journal}{Journal of Fluid Mechanics} \bibinfo{volume}{835},
  \bibinfo{pages}{252--270}.
\bibitem[{Tolley et~al.(2014)Tolley, Shepherd, Mosadegh, Galloway, Wehner,
  Karpelson, Wood and Whitesides}]{tolley2014resilient}
\bibinfo{author}{Tolley, M.T.}, \bibinfo{author}{Shepherd, R.F.},
  \bibinfo{author}{Mosadegh, B.}, \bibinfo{author}{Galloway, K.C.},
  \bibinfo{author}{Wehner, M.}, \bibinfo{author}{Karpelson, M.},
  \bibinfo{author}{Wood, R.J.}, \bibinfo{author}{Whitesides, G.M.},
  \bibinfo{year}{2014}.
\newblock \bibinfo{title}{A resilient, untethered soft robot}.
\newblock \bibinfo{journal}{Soft robotics} \bibinfo{volume}{1},
  \bibinfo{pages}{213--223}.
\bibitem[{Yu et~al.(2012)Yu, Ding, Yang, Tan, Wang and Zhang}]{yu2012bio}
\bibinfo{author}{Yu, J.}, \bibinfo{author}{Ding, R.}, \bibinfo{author}{Yang,
  Q.}, \bibinfo{author}{Tan, M.}, \bibinfo{author}{Wang, W.},
  \bibinfo{author}{Zhang, J.}, \bibinfo{year}{2012}.
\newblock \bibinfo{title}{On a bio-inspired amphibious robot capable of
  multimodal motion}.
\newblock \bibinfo{journal}{IEEE/ASME Transactions On Mechatronics}
  \bibinfo{volume}{17}, \bibinfo{pages}{847--856}.
\bibitem[{Zhang and Goldman(2014)}]{zhang2014effectiveness}
\bibinfo{author}{Zhang, T.}, \bibinfo{author}{Goldman, D.I.},
  \bibinfo{year}{2014}.
\newblock \bibinfo{title}{The effectiveness of resistive force theory in
  granular locomotion}.
\newblock \bibinfo{journal}{Physics of Fluids} \bibinfo{volume}{26},
  \bibinfo{pages}{101308}.

\end{thebibliography}
\end{document}